\documentclass[journal]{IAENGtran}
\ifCLASSINFOpdf
   \usepackage[pdftex]{graphicx}
   \DeclareGraphicsExtensions{.pdf,.jpeg,.png}
\else
   \usepackage[dvips]{graphicx}
   \DeclareGraphicsExtensions{.eps}
\fi
\usepackage{algorithm}
\usepackage{algpseudocode}

\usepackage{multirow}
\usepackage{makecell}
\usepackage{pgfplots}
\usepackage{filecontents}

\usepackage{array}
\newcolumntype{M}[1]{>{\centering\arraybackslash}m{#1}}

\begin{document}
%
\title{A Hybrid Approach for Improved Low Resource Neural Machine Translation using Monolingual Data}
%
%
%

\author{Idris~Abdulmumin*,~\IAENGmembership{Member,~IAENG,}
        Bashir~Shehu~Galadanci,
        Abubakar~Isa,
        Habeebah~Adamu~Kakudi,
        and~Ismaila~Idris~Sinan
\thanks{Manuscript received February 09, 2021; revised September 13, 2021. 
This work is supported by the National Information Technology Development Agency under the National Information Technology Development Fund PhD Scholarship Scheme 2018.}
\thanks{*Corresponding Author. Idris Abdulmumin is a PhD Candidate at the Department of Computer Science, Bayero University, Kano, Nigeria, on a study leave from Ahmadu Bello University, Zaria, Kaduna, Nigeria, (phone: +234(0)-806-295-5509; e-mail: iabdulmumin@abu.edu.ng).}%
\thanks{Bashir Shehu Galadanci is an Associate Professor at the Department of Software Engineering, Bayero University, Kano, Kano, Nigeria, (e-mail: bsgaladanci.se@buk.edu.ng).}%
\thanks{Abubakar Isa is a Lecturer at the Department of Computer Science, Ahmadu Bello University, Zaria, Kaduna, Nigeria, (e-mail: abubakarisa@abu.edu.ng).}%
\thanks{Habiba Adamu Kakudi is a Senior Lecturer and Head of Department at the Department of Computer Science, Bayero University, Kano, Kano, Nigeria, (e-mail: hakakudi.cs@buk.edu.ng).}%
\thanks{I. I. Sinan is a PhD Candidate at the Africa Center of Excellence on Technology Enhanced Learning, National Open University of Nigeria, Nigeria, (e-mail: isinan@noun.edu.ng).}}

\maketitle

\pagestyle{empty}
\thispagestyle{empty}

\begin{abstract}
Many language pairs are low resourced, meaning the amount and/or quality of the available parallel data between them is not sufficient to train a neural machine translation (NMT) model which can reach an acceptable standard of accuracy. Many works have explored using the readily available monolingual data in either or both of the languages to improve the standard of translation models in low, and even high, resource languages. One of the most successful of such works is the back-translation that utilizes the translations of the target language monolingual data to increase the amount of the training data. The quality of the backward model which is trained on the available -- often low resource -- parallel data has been shown to determine the performance of this approach. Despite this, the standard back-translation is meant only to improve the performance of the forward model on the available monolingual target data. A previous study proposed an iterative back-translation approach that, unlike in traditional back-translation, relies on both the target and source monolingual data to improve the two models over several iterations. In this work, however, we proposed a novel approach that enables both of the backward and forward models to benefit from the monolingual target data through a hybrid of self-learning and back-translation respectively. Experimental results have shown the superiority of the proposed approach over the traditional back-translation method on English-German low resource neural machine translation. We also proposed an iterative self-learning approach that outperforms the iterative back-translation while also relying only on the monolingual target data and requiring the training of less models.
\end{abstract}

\begin{IAENGkeywords}
Self-Learning, Back-Translation, Iterative Self-Learning, Iterative Back-Translation, Neural Machine Translation, Natural Language Processing
\end{IAENGkeywords}

%
\IAENGpeerreviewmaketitle

\section{Introduction}
%
%
%
%

\IAENGPARstart{T}{he} Neural Machine Translation (NMT) \cite{Bahdanau2014,Gehring2017,Vaswani2017} has been used to model state-of-the-art translation systems for many high-resource languages \cite{Edunov2018,Hassan2018,Hoang2018,Ott2018}. These systems are used for translations of day-to-day conversations, documents and web contents or as part of other cross-lingual studies such as in sentiment analysis \cite{Abdalla2017} or information retrieval \cite{Bracewell2008}, among others. For many language pairs though, the amount and/or quality of parallel data is not enough to train an NMT model whose accuracy can reach an acceptable standard \cite{Edunov2018,Zoph2016}. This category of language pairs is known as low resource. Many works have explored how to use of either the easier-to-get monolingual data through self-training \cite{Specia2018,Ueffing2006}, forward translation \cite{Zhang2016}, language modelling \cite{Lample2019,Gulcehre2017} and back-translation \cite{Graca2019,Poncelas2018,Poncelas2019,Poncelas2019a} or other intermediate language \cite{Adusumilli2007} or languages \cite{Zoph2016} to improve the quality of translation models in this category of languages -- and even high resource languages.

The back-translation has so far been one of the most successful methods \cite{Edunov2018,Lample2019,Lioutas2020}, involving the use of the translations of the target language monolingual data to increase the amount of the training data \cite{Sennrich2016a}. The additional parallel data consists of authentic sentences in the target language and their translations -- synthetic sentences in the source language -- generated using a reverse (backward) model that is trained on the available parallel data -- see the procedure in Algorithm~\textbf{\ref{algo-bt}}. The approach has proven to be successful at improving the quality of translations in high, middle and low resourced languages \cite{Edunov2018,Burlot2018}. Many studies have shown that the quality of the backward system influences the performance of the ultimate NMT model \cite{Hoang2018,Sennrich2016a,Cotterell2018}. In low resource conditions, the available parallel data may not be able to train a standard backward model and the quality of the additional data generated using this model may hurt the quality of the final model. Despite this, the aim of standard back-translation has always been to improve the performance of the target NMT model by providing sufficient training data.

Some previous works have proposed various methods to improve the performance of the backward model during training. These methods include iterative back-translation \cite{Hoang2018,Cotterell2018}, transfer learning \cite{Dabre2019,Kocmi2019,Luo2020}, self-training \cite{Abdulmumin2021} and the training of a bi-directional translation model for both backward and forward translations \cite{Niu2018}. Others have tried to mask the deficiencies of the backward model either during inference by generating multiple translations of the same target sentence using sampling to average-out the errors in individual translations \cite{Imamura2018} and noising the output of beam search \cite{Sennrich2016}; or reducing the effects of the errors in the synthetic data when training the forward model through methods such as tagged back-translation \cite{Caswell2019,Yang2019} and synthetic data pre-training \cite{Abdulmumin2019a}.

In this work, we present a hybrid approach that utilizes the monolingual target data to improve both the forward and backward models in back-translation. In this approach, we used the synthetic data to enhance the backward model through self-learning and the standard back-translation for improving the forward model. The approach was preliminary investigated in \cite{Abdulmumin2021} and it was shown to achieve positive results. Earlier use of stand-alone self-training in machine translation proposed extra methods of either using quality estimation \cite{Specia2018} or freezing of the decoder weights \cite{Zhang2016} when training on the synthetic side of the training data. It was suggested that the mistakes in the synthetic data will hurt the performance of the self-trained model \cite{Specia2018,Zhang2016}. Instead, \cite{Abdulmumin2021} showed that self-training is capable of improving the quality of the backward model even without using either of the specialized approaches. It was shown that using all of the synthetic data generated by the backward model to help in re-training the backward model improved its performance. The work, though, did not show the benefits or otherwise of using any of the specialized approach in cleaning the data, especially in low resource languages. It also did not investigate if the model can continue to learn from its output through an iterative process.

\begin{algorithm}[t!]
\renewcommand{\arraystretch}{1.3}
\caption{Standard Back-translation \cite{Sennrich2016}}
\label{algo-bt}
\textbf{Input}: Parallel data, \(D^P =\{(x^{(u)}, y^{(u)})\}_{u=1}^U\), and Monolingual target data, \(Y =\{(y^{(v)})\}_{v=1}^V\)
\begin{algorithmic}[1]
\Procedure{Standard Back-Translation}{}
	\State Train backward model \(M_{x \leftarrow y}\) on bilingual data \(D^P\);
	\State Let \(D^\ast = \) synthetic parallel data generated for \(Y\) using \(M_{x \leftarrow y}\);
	\State Train forward model \(M_{x \rightarrow y}\) on bilingual data \(D^P \cup D^\ast\);
\EndProcedure
\end{algorithmic}
\textbf{Output}: \(M_{x \leftarrow y}\) and improved \(M_{x \rightarrow y}\) models
\end{algorithm}

This work, therefore, investigates the effects of synthetic data cleaning using automatic quality estimation when training the backward model. We observed that while self-learning may improve the backward model using all of the available synthetic data, selecting only a subset of the data may result in a superior but less generic model. We then investigated the use of iterative self-training with quality estimation as proposed in \cite{Specia2018}, enabling the backward model to be trained on all the monolingual data -- enhancing its generalization ability. For low resource languages, readily available quality estimation systems or the data to train such systems may not be available. This may limit the implementation of the approach.  We, therefore, proposed a novel iterative approach that relies only on all the available monolingual target data to improve the backward model before finally generating a much improved synthetic data for the forward model's training. Experimental results show that our approach is superior to the standard back-translation and the approach proposed in \cite{Abdulmumin2021}; and that our iterative approach is superior to the iterative back-translation \cite{Hoang2018} while also requiring less number of models to be trained.

We thus make the following contributions in this paper:
\renewcommand{\labelitemi}{\textbullet}
\begin{itemize}
\item we showed that although the self-training approach improves the performance of backward model without data cleaning and/or freezing learned parameters, selecting the synthetic data with the best quality results in a better self-trained backward model;
\item we showed that selecting the best synthetic data and using it for the self-training is just enough to train better backward model;
\item we proposed a novel back-translation approach that implements an iterative self-learning approach to improve the performance of the backward model on the monolingual target data in low resource neural machine translation systems;
\item we showed that our iterative self-training approach for training the backward model relies only the monolingual target data unlike in previous iteration in back-translation approaches that also require the monolingual source data;
\item we compared two variants of the iterative self-learning approaches: with \cite{Specia2018} and without quality estimation, and found both to be capable of further improving the quality of the backward model;
\item we found that the iterative self-learning without quality estimation can reach a performance that is superior to that which uses the quality estimation while also being faster;
\item we showed that the improvements achieved in the performance of the backward models translate into better forward models; and
\item experimental results on English-German low resource translation indicate that the method is applicable on, and can be generalized to, other low resource languages.
\end{itemize}

The remainder of this paper is organized as follows: In Section~\ref{sec:literature}, we reviewed the related works. We presented the proposed methods in Section~\ref{sec:methods}. We presented the experimental set-up including the choice of NMT architecture and the dataset used in Section~\ref{sec:experiment}. We discussed the experiments conducted and the results obtained; and the findings of the research work in Sections~\ref{sec:results} and \ref{sec:discuss} respectively and, finally, the paper was concluded and directions for future work were proposed in Section~\ref{sec:conclude}.

\section{Related Works}
\label{sec:literature}
This section presents prior work on back-translation, iterative back-translation, forward translation and self-training and automatic quality estimation for self-training.

\subsection{Back-Translation}

Back-translation is an approach that enables augmenting the available parallel data with the back-translations of monolingual sentences in the target language to increase the amount of data for training improved translation models. In NMT, \cite{Sennrich2016a} proposed using this approach to solve the problem that the architecture struggles with – lack of adequate training data. The approach has since been used for training state-of-the-art translation models for many language pairs \cite{Edunov2018,Hassan2018}. The success of the approach has been shown to rely on three factors: the quality of the synthetic data which depends on the quality of the backward model \cite{Hoang2018,Graca2019,Imamura2018,Sennrich2016,Yang2019} and the synthetic data generation method used \cite{Edunov2018,Graca2019,Sennrich2016a,Imamura2018}; the ratio of the synthetic to authentic parallel data used \cite{Poncelas2018,Fadaee2018}; and the architecture of the NMT used \cite{Sennrich2016a}.

Many studies have shown that the quality of the backward system influences the performance of the ultimate NMT model more than other factors \cite{Hoang2018,Sennrich2016a,Cotterell2018}. In low resource conditions, the available parallel data may not be able to train a standard backward model and the quality of the additional data generated using this model may hurt the quality of the final model. Various research works have proposed methods aimed at training a standard backward model no matter the amount of authentic data available and, therefore, improving the quality of the synthetic data. \cite{Dabre2019} and \cite{Kocmi2019} investigated pre-training the backward model on a high-resource language pair (parent) and fine-tuning the model on the low resource language pair (child) through transfer learning \cite{Zoph2016,Kocmi2018}. \cite{Graca2019} posits that selecting the appropriate synthetic data generation method can offset the deficiencies of the backward model. \cite{Abdulmumin2021} investigated the application of the self-training approach to improve the backward model in low resource NMT.

\subsection{Iterative Back-Translation}

Back-translation relies on the parallel data between the two languages and the target monolingual data to improve the forward model only. The iterative back-translation, on the other hand, relies on the monolingual source and target data to improve the backward and forward models respectively. The backward model generates synthetic data to improve the forward model and the forward model does the same for the backward model. The process is repeated until a set amount of iterations or desired quality is reached.

The iterative back-translation was proposed in the work of \cite{Hoang2018} to enable the backward and forward models to be improved over several iterations. The work hypothesized that if the back-translation approach has shown to be successful at improving the standard of machine translation models, repeating the procedure will further enhance the quality of the translations produced. Whereas the synthetic data generated by the backward model is mainly used to enhance the forward model in standard back-translation, iterative back-translation enables the backward model, also, to be improved on the synthetic data generated by the forward model. The approach has been shown to successfully improve both the backward and forward models \cite{Hoang2018,Cotterell2018,Caswell2019,Yang2019}.

Standard back-translation is used throughout the process but it relies also on the monolingual source data, deviating from the requirements of traditional back-translation. Also, where the monolingual source data is not available, the procedure cannot be applied. The approach, implemented with standard back-translation, was shown to reduce the performance of English-Romanian machine translation, improving the performance only after applying a tagging approach \cite{Caswell2019}.

\subsection{Forward Translation and Self-Training or Self-Learning}
\label{lit:sl}

Self-learning is an approach where a model learns from its own output. In machine translation, a model is trained on the available parallel data and is then used to translate a given set of monolingual source sentences to generate a pseudo-parallel (synthetic) training data. This data is then used to train a better model than the generating model. Ueffing \cite{Ueffing2006} first used the approach to improve phrased-based statistical machine translation systems. The work uses an existing system to generate the translations of a new set of source data. The confidence score of each of the translated sentences is estimated and based on these scores, translations that are considered reliable are extracted and used as additional training data for improving the existing system.

In neural machine translation, \cite{Zhang2016} explored the use of monolingual source data to improve the accuracy of a translation model through forward translation. A baseline NMT model was trained and used to generate synthetic parallel data -- authentic source + synthetic target data. Since the NMT architecture consists of an encoder-decoder architecture and the encoder is trained on the source data while the decoder learns representation based on the representations of the encoder and the target data, the use of synthetic data -- which often contains mistakes -- to train any or both of the encoder and decoder may deteriorate their performance. To mitigate this problem, the authors proposed freezing the parameters of the decoder when training on the synthetic target data. By this, the useful representations learned on the authentic data will not be unlearned when training on the additional data.

In an approach similar to \cite{Ueffing2006}, \cite{Specia2018} used self-training on phrased-based and rule-based statistical machine translation systems to better their performance. Their work investigated the use of iterative self-training with quality estimation in a process that utilizes all of the monolingual data to improve the generating models. First, a baseline translation model is trained and is used to translate a given set of monolingual sentences. The quality of the synthetic (translated) sentences are determined using a quality estimation system. The best sentences are selected and added to the training data. This new (bigger) training data is used to retrain the translation model. the rest of the sentences whose translations were not selected are translated again using the improved model and the procedure continues until all of the monolingual sentences are exhausted.

The work of \cite{Abdulmumin2021} investigated the feasibility of using the self-learning approach only -- without quality estimation and/or freezing of any parameters -- to train a better NMT model. It was determined that not only is it simpler than the other methods, it is also capable of improving baseline translation models. The authors used the approach to enhance the performance of a backward model in the back-translation approach. The quality of the model was improved, resulting in a better target NMT model. The work, though, only showed the applicability of the self-training approach but did not investigate the extent -- through iterative training -- to which the backward model will continue to be improved if it is trained on the synthetic data it generates. It also did not show the benefits or otherwise of data selection on the quality of the backward model. This work, therefore, proposes the iterative self-learning approach to improve the quality of the backward model. It also extends the work in \cite{Abdulmumin2021} to compare the use or otherwise of quality estimation in determining the best synthetic data for self-training the backward model.

\subsection{Automatic Quality Estimation for Self-Training}
Quality Estimation (QE) is a method used for Translation Quality Assessment. The method was introduced to help in determining the quality of translations when reference (human-generated) text is not available \cite{Chen2017}. When such texts are available, traditional metrics such as the Bi-Lingual Evaluation Understudy (BLEU) \cite{Papineni2002}, the METEOR \cite{Lavie2009} and the Z-numbers-influenced quality estimation metric \cite{Qiu2020} are used. The technique is particularly useful for indicating the reliability of translations especially to users who cannot read the source language \cite{Specia2018}, for deciding the translations that are fit for publishing and those requiring human or automatic revision (post-editing) \cite{Martins2017}. This technique has been shown to reduce the time taken for post-editing \cite{Specia2013}.

Work on the technique was started in early 2000s \cite{Specia2018} with works such as \cite{Ueffing2003,Blatz2004} and has been gaining prominence especially with the ever increasing reliance on machine translation models. Recent works such as \cite{Chen2017,Martins2017,Specia2013,Kim2017,Kepler2019,Wang2018} have achieved tremendous improvements in translation quality estimation, mostly using neural models to train better systems that are capable of estimating the quality of translations produced.

The QE technique was used in self-learning in works such as \cite{Specia2018,Ueffing2006} to determine the translations that are considered fit for retraining the generating model and has been shown to improve the performance of statistical machine translation systems. Instead of relying on the translated sentences, which can be good or bad, the QE system enables the training of better translation models only on data that is estimated to be good.

\section{Methodology}
\label{sec:methods}
In this section, we described the neural architecture used in building the models. We also described the three self-learning methods proposed for improving the performance of the back-translation approach.

\subsection{Neural Machine Translation (NMT)}

The NMT is simpler than other forms of machine translation architectures because an NMT model can learn the probability of mapping input sentences in the source language to the target sentences in another language by relying only on the training parallel data between the two languages \cite{Yang2019}. The NMT architecture is a sequence-to-sequence system made of encoder and decoder neural networks that models the conditional probability of a target sentence given a source sentence \cite{Bahdanau2014,Luong2015,Sutskever2014}. The encoder converts the sentences in the source language into a set of vectors while the decoder converts the set of vectors, one word at a time, into equivalent sentences in the target language through an attention mechanism. The attention mechanism was introduced to keep track of context in longer sentences \cite{Bahdanau2014}. Our work is based on a unidirectional LSTM \cite{Hochreiter1997} encoder-decoder architecture with Luong attention \cite{Luong2015}, a recurrent neural network RNMT architecture. It can be implemented using other architectures such as the Transformer \cite{Vaswani2017,Dehghani2019} and the convolutional neural network NMT (CNMT) \cite{Gehring2017,Wu2019}.

The NMT model produces the translation sentence by generating one target word at every time step. Given an input sequence \(X=(x_1,...,x_{T_x})\) and previously translated words \((y_1,...,y_{i-1})\), the probability of the next word \(y_i\) is \begin{equation} p(y_i|y_1,...,y_{i-1},X) = g(y_{i-1},s_i,c_i) \end{equation} where \(s_i\) is the decoder hidden state for time step \(i\) and is computed as \begin{equation} s_i = f(s_{i-1},y_{i-1},c_i). \end{equation} Here, \(f \) and \(g\) are nonlinear transform functions, which can be implemented as long short-term memory (LSTM) network \cite{Hochreiter1997} or gated recurrent units (GRU) \cite{Cho2014} in recurrent neural machine translation (RNMT), and \(c_i\) is a distinct context vector at time step \(i\), which is calculated as a weighted sum of the input annotations \(h_j\), , which is a concatenation of the forward and backward hidden states \(\overrightarrow{h_j}\) and \(\overleftarrow{h_j}\) respectively.
\begin{equation}
c_i=\sum_{j=1}^{T_x} a_{i,j}h_j
\end{equation}
where \(h_j\) is the annotation of \(x_j\) calculated by a bidirectional Recurrent Neural Network. The weight \(a_{i,j}\) for \(h_j\) is calculated as
\begin{equation}
a_{i,j} = \frac{\exp{e_{i,j}}}{\sum_{t=1}^{T_x} \exp{e_{i,t}}}
\end{equation}
and
\begin{equation}
e_{i,j} = v_a\tanh(Ws_{i-1}+Uh_j)
\end{equation}
where \(v_a\) is the weight vector, \(W\) and \(U\) are the weight matrices.

All of the parameters in the NMT model, represented as $\theta$, are optimized to maximize the following conditional log-likelihood of the \(M\) sentence aligned bilingual samples
\begin{equation}
L(\theta) = \frac{1}{M}\sum_{m=1}^{M} \sum_{i=1}^{T_y}\log(p(y_i^m|y_{<i}^m, X^m,\theta))
\end{equation}

\subsection{Overview of the Proposed Methods}
This section presents the self-learning and iterative self-learning both with and without quality estimation.

\begin{algorithm}[h]
\renewcommand{\arraystretch}{1.3}
\caption{Self-Learning Enhanced Back-translation for Low Resource NMT \cite{Abdulmumin2021}}
\label{algo-slbt}
\textbf{Input}: Parallel data, \(D^P =\{(x^{(u)}, y^{(u)})\}_{u=1}^U\), Monolingual target data, \(Y =\{(y^{(v)})\}_{v=1}^V\) and \(n = \) number of required synthetic data
\begin{algorithmic}[1]
\Procedure{Self-Learning}{}
    \State Train backward model \(M_{x \leftarrow y}\) on bilingual data \(D^P\);
    \State Use \(M_{x \leftarrow y}\) to generate synthetic data, \(D^\prime = \{(x^{(v)}, y^{(v)})\}_{v=1}^V\) for \(y \in Y\);
    \State Train improved backward model \(M_{x \leftarrow y}^+\) on bilingual data \(D^P\) and \(D^\prime\);
\EndProcedure
\Procedure{Improved Back-Translation}{}
    \State Let \(D^s\) = synthetic parallel corpora generated for \(Y\) using \(M_{x \leftarrow y}^+\);
    \State Train forward model \(M_{x \rightarrow y}\) on bilingual data \(D^P\) and \(D^s\);
\EndProcedure
\end{algorithmic}
\textbf{Output}: improved backward and forward models, \(M_{x \leftarrow y}\) and \(M_{x \rightarrow y}\) respectively
\end{algorithm}

\subsubsection{Enhanced Back-Translation using Self-Learning}
\label{sec:sl}
The approach was proposed and preliminarily investigated in \cite{Abdulmumin2021} to enhance the performance of the synthetic data generating model. As presented in Algorithm~\textbf{\ref{algo-slbt}}, given the authentic parallel data, \(D^P\), a backward model, \(M_{x \leftarrow y}\), is trained. This model is used to translate the available monolingual data to generate the synthetic parallel data, \(D^\prime\). Finally, \(D^\prime\) is used with \(D^P\) to train an improved backward model \(M_{x \leftarrow y}^+\) through the pre-training and fine-tuning \cite{Abdulmumin2019a} technique. The technique has been shown to outperform other methods such as the standard forward translation, tagged forward translation (similar to the tagged back-translation \cite{Caswell2019}) both during self-training and back-translation \cite{Abdulmumin2021,Abdulmumin2019a}.

\subsubsection{Enhanced Back-Translation using Self-Learning with Quality Estimation}
\label{sec:slqe}

We first extended the work in Section~\ref{sec:sl} to find the benefits or otherwise of using a quality estimation system to the training of the backward model. This approach is illustrated in Algorithm~\textbf{\ref{algo-slqebt}}. Instead of using the whole of the synthetic data as in \cite{Abdulmumin2021}, we used a Quality Estimation (QE) system to select the best among the synthetic data. This can be one-third, two-thirds or any fraction of the synthetic data. The best synthetic data is then used with the authentic data to enable the training of an improved backward. The quality estimation was deployed to ensure that only the best synthetic data is used in the approach. The implementation of this approach might, therefore, be presented with two issues. First, low resource languages may not have a readily available quality estimation system to enable the selection of the training synthetic data. Secondly, since only a subset of the monolingual data was used, the vocabulary of the translation model will, therefore, be limited and the model, although superior, will not be as generic as a model trained on the whole of the synthetic data. This trade-off between quality and generalization may need to be reduced.

\begin{algorithm}[h]
\renewcommand{\arraystretch}{1.3}
\caption{Self-Learning with Quality Estimation Enhanced Back-translation for Low Resource NMT}
\label{algo-slqebt}
\textbf{Input}: Parallel data, \(D^P =\{(x^{(u)}, y^{(u)})\}_{u=1}^U\), Monolingual target data, \(Y =\{(y^{(v)})\}_{v=1}^V\) and \(n = \) number of required synthetic data
\begin{algorithmic}[1]
\Procedure{Self-Learning with Quality Estimation}{}
    \State Train backward model \(M_{x \leftarrow y}\) on bilingual data \(D^P\);
    \State Use \(M_{x \leftarrow y}\) to generate synthetic data, \(D^\prime = \{(x^{(v)}, y^{(v)})\}_{v=1}^V\) for \(y \in Y\);
    \State Using a \(QE\) system, select best \(n\)synthetic data, \(D^n\), from \(D^\prime\);
    \State Train improved backward model \(M_{x \leftarrow y}^+\) on bilingual data \(D^P\) and \(D^n\);
\EndProcedure
\Procedure{Improved Back-Translation}{}
    \State Let \(D^s\) = synthetic parallel corpora generated for \(Y\) using \(M_{x \leftarrow y}^+\);
    \State Train forward model \(M_{x \rightarrow y}\) on bilingual data \(D^P\) and \(D^s\);
\EndProcedure
\end{algorithmic}
\textbf{Output}: improved backward and forward models, \(M_{x \leftarrow y}\) and \(M_{x \rightarrow y}\) respectively
\end{algorithm}

As presented in Agorithm~\textbf{\ref{algo-slqebt}}, given the authentic parallel data, \(D^P\), a backward model, \(M_{x \leftarrow y}\), is trained. This model is used to translate the available monolingual data to generate the synthetic parallel data, \(D^\prime\). A \(QE\) system is then used to select the best \(n\) synthetic data, \(D^n\). Finally, \(D^n\) is used with \(D^P\) to train an improved backward model \(M_{x \leftarrow y}^+\).


\begin{algorithm}[h]
\renewcommand{\arraystretch}{1.3}
\caption{Iterative Self-Learning with Quality Estimation Enhanced Back-translation for Low Resource NMT}
\label{algo-islqebt}
\textbf{Input}: Parallel data, \(D^P =\{(x^{(u)}, y^{(u)})\}_{u=1}^U\), Monolingual target data, \(Y =\{(y^{(v)})\}_{v=1}^V\) and \(n = \) number of required synthetic data
\begin{algorithmic}[1]
\Procedure{Iterative Self-Learning with Quality Estimation}{}
    \State Let \(D = D^P\)
    \Repeat
        \State Train backward model \(M_{x \leftarrow y}\) on bilingual data \(D\);
        \State Use \(M_{x \leftarrow y}\) to generate synthetic data, \(D^\prime = \{(x^{(v)}, y^{(v)})\}_{v=1}^V\) for \(y \in Y\);
        \State Using a \(QE\) system, select best \(n\) synthetic data, \(D^n\), from \(D^\prime\);
        \State Remove \(Y^n = \) selected monolingual sentences from \(Y\);
        \State Let \(D = D \cup D^n\);
    \Until{monolingual data is exhausted;}
\EndProcedure
\Procedure{Improved Back-Translation}{}
    \State Let \(D^s\) = synthetic parallel corpora generated for \(Y\) using improved \(M_{x \leftarrow y}\);
    \State Train forward model \(M_{x \rightarrow y}\) on bilingual data \(D^P\) and \(D^s\);
\EndProcedure
\end{algorithmic}
\textbf{Output}: improved backward and forward models, \(M_{x \leftarrow y}\) and \(M_{x \rightarrow y}\) respectively
\end{algorithm}

\subsubsection{Enhanced Back-Translation using Iterative Self-Learning with Quality Estimation}
\label{sec:islqe}

In this approach we aim to expand the utilization of the monolingual data and enhance the capability of the backward model to generalize more on unseen data by learning on all the available monolingual data. We, therefore, used the iterative self-learning approach as proposed in \cite{Specia2018}. The approach, as explained in Section~\ref{lit:sl}, was successfully applied to improve the performance of statistical machine translation models using all of the available monolingual data over several iterations. At each iteration, the best of the remaining synthetic data is used to self-train the backward model.

As illustrated in Algorithm~\textbf{\ref{algo-islqebt}}, given the parallel data \(D^P\), we train a backward model \(M_{x \leftarrow y}\). This model is used to generate the synthetic data \(D^\prime\) by translating the available monolingual data, \(Y\). We then used a Quality Estimation model to estimate the quality of the generated synthetic data. The top \(n\) synthetic parallel data are removed and added to the training data to train an improved backward model. This stage is labeled as iteration \(0\). This stage is the same as the self-training with quality estimation described in Section~\ref{sec:slqe}. The process is repeated until all the available monolingual data is exhausted, using the improved backward model to generate the synthetic data at each iteration. Afterwards, the final improved backward model is used to generate the synthetic data for training the forward model through back-translation.

\begin{figure*}
\centering
\includegraphics[clip, trim=2.5cm 17.3cm 6.0cm 3.0cm, width=0.9\textwidth]{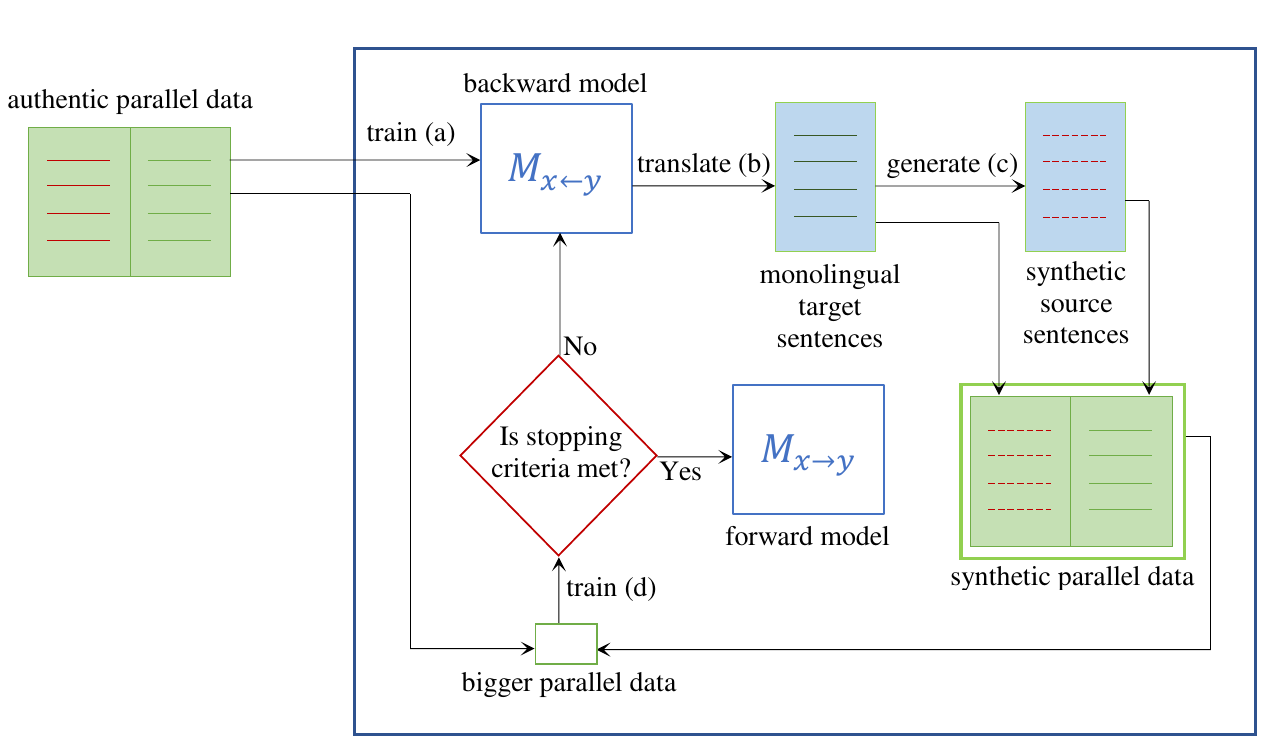}
\caption{Iterative Self-Learning: Enabling the backward model to learn iteratively on its output -- the synthetic data}
\label{fig:isl}
\end{figure*}

\begin{algorithm}[t]
\renewcommand{\arraystretch}{1.3}
\caption{Iterative Self-Learning Enhanced Back-translation for Low Resource NMT}
\label{algo-islbt}
\textbf{Input}: Parallel data, \(D^P =\{(x^{(u)}, y^{(u)})\}_{u=1}^U\), and Monolingual target data, \(Y =\{(y^{(v)})\}_{v=1}^V\)
\begin{algorithmic}[1]
\Procedure{Iterative Self-Learning}{}
    \State Let \(D = D^P\)
    \Repeat
        \State Train backward model \(M_{x \leftarrow y}\) on bilingual data \(D\);
        \State Use \(M_{x \leftarrow y}\) to generate synthetic data, \(D^\prime = \{(x^{(v)}, y^{(v)})\}_{v=1}^V\) for \(y \in Y\);
        \State Let \(D = D^P \cup D^n\);
    \Until{convergence condition is reached;}
\EndProcedure
\Procedure{Improved Back-Translation}{}
    \State Let \(D^s\) = synthetic parallel corpora generated for \(Y\) using improved \(M_{x \leftarrow y}\);
    \State Train forward model \(M_{x \rightarrow y}\) on bilingual data \(D^P\) and \(D^s\);
\EndProcedure
\end{algorithmic}
\textbf{Output}: improved backward and forward models, \(M_{x \leftarrow y}\) and \(M_{x \rightarrow y}\) respectively
\end{algorithm}

\subsubsection{Iterative Self-Learning Enhanced Back-Translation}
\label{sec:isl}

The approach in \cite{Abdulmumin2021} is a one-off utilization of self-learning to enhance the capability of the backward model to produce improved translations. The extension in Section~\ref{sec:slqe} is expected to improve the model's performance since the quality of the synthetic data will be better. But the approach is expected to be limited in terms of its generalization ability. The approach presented in Section~\ref{sec:islqe} is expected to achieve a better performance because it combines the best of both worlds -- iterative self-training and quality estimation. But the approach is also limited in its reliance and trust on a quality estimation system to achieve the desired result. A poor quality estimation model, the absence altogether of such models or the lack of data to train them will limit the application of this approach on many languages. Also, the synthetic parallel data that are adjudged by the quality estimation system are considered final and, therefore, will not benefit from the improved backward model in subsequent iterations even though re-translating such data may have led to improving their quality even further.

Having shown the limitations of previous approaches, we proposed a novel approach for using the synthetic data to iteratively self-train the backward model, indirectly improving the forward model in back-translation. The method is a hybrid of iterative self-learning and back-translation with each method using the monolingual data to train better models. The main essence for the hybridization is to utilize the capabilities of the back-translation approach to generate additional data for training improved translation models and the iterative self-learning approach to improve the data-generating (backward) model. In this approach, improved backward models -- obtained after self-training -- are used for generating better versions of the synthetic data for the next iteration. After reaching the desired number of iterations or translation quality, the final backward model is used to generate the synthetic data for training the forward model.

We, therefore, implemented the iterative approach that enables the generation of a better synthetic data while utilizing all of the monolingual data, see Algorithm~\textbf{\ref{algo-islbt}}. Similar to the approach in \cite{Abdulmumin2021}, we use the parallel data \(D^P\) to train a backward model. This model is used to generate the synthetic data \(D^\prime\) by translating the available monolingual data, \(Y\). In this method, the improved synthetic data is not used, yet, with the authentic parallel data to train the forward model. Instead, the process is labeled as iteration \(0\). \(D^\prime\) is used with the authentic data \(D^P\) to train a better backward model for the first iteration – iteration \(1\). This process is repeated until the condition(s) of convergence are reached. This condition(s) can be set as a fixed number of iterations or until after the performance of subsequent backward models is not improved or not more than a set threshold over some number of evaluations. Afterwards, the final improved backward model is used to generate the synthetic data for training the forward model through back-translation. Fig.~\ref{fig:isl} better illustrate this approach.

In the iterative back-translation approach, the number, \(n(k)\), of models – backward and forward – required to be trained for \(k\) iterations is
\begin{equation}
n(k)=2k+2.
\end{equation}
This means \(k+1\) forward and backward models each. For all the iterative self-learning enhanced back-translation we proposed in this work, the number, \(n(k)\), of models required to be trained for \(k\) iterations is
\begin{equation}
n(k)=k+3.
\end{equation}
Therefore, it can be seen from equations (1) and (2) that for the same number of iterations, \(k\), the number of models required to be trained using the iterative self-learning enhanced back-translation is \((k-1)\)-less than that of the iterative back-translation. In this approach, we train \(k+2\) backward models but only one forward model -- the target forward model. For \(k = 0\), the approach is the self-trained enhanced back-translation and we have 3 models, while for \(k > 1\), it becomes the iterative self-learning enhanced back-translation.

\section{Experimental Set-up}
\label{sec:experiment}

In this section, we presented the various the experimental set-up and the dataset that were used to implement the proposed and current approaches for evaluation. The models that were trained -- including the baseline, the backward model before and after applying the various self-training approaches -- were also explained.

\subsection{Set-up}
\label{sec:setup}
In this work, we followed the same set-up as \cite{Abdulmumin2021}. We used the TensorFlow \cite{Abadi2015} implementation of OpenNMT -- OpenNMT-tf \cite{Klein2017}. We used the NMTSmallV1 configuration which is a 2-layer unidirectional LSTM encoder-decoder configuration with 512 hidden units each and Luong attention \cite{Luong2015}. For both the source and target languages, we used a vocabulary size of 50,000. We used Adam optimizer \cite{Kingma2015}, a training batch size of 64, a dropout probability of 0.3 and a static learning rate of 0.0002.

The works of \cite{Abdulmumin2021} and \cite{Abdulmumin2019a} have shown that a model trained on a mixture of synthetic and authentic data benefits most from pre-training on the synthetic data and fine-tuning on the authentic data than using other approaches such as noising \cite{Edunov2018} and tagging \cite{Caswell2019}. We, therefore, adopted the pre-training and fine-tuning approach to train the enhanced backward model.

For quality estimation and to implement the work of \cite{Specia2018}, we used a pre-trained Predictor-Estimator \cite{Kim2017} model provided in the OpenKiwi’s \cite{Kepler2019} tutorial for estimating the quality of the synthetic sentences. The model is based on a Recurrent Neural Network architecture that utilizes multi-level task learning for determining the quality of synthetic sentences. It is built on a neural predictor model and another neural quality estimation model learnt jointly using a stack propagation method \cite{Zhang2016a}.

\begin{table}[h]
\renewcommand{\arraystretch}{1.3}
\caption{Data Statistics}
\label{tab1}
\begin{tabular}{|c||c||c||c|}
\hline
\multirow{2}{*}{data} & \multicolumn{3}{c|}{train} \\ \cline{2-4}
 & sentences & \multicolumn{2}{c|}{words (vocab)} \\ \hline
\multirow{2}{*}{} & \multirow{2}{*}{} & En & De \\ \cline{3-4}
 IWSLT’14 En-De & 153, 348 & \makecell[tc]{2,706,255\\ (54,169)} & \makecell[tc]{3,311,508\\ (25,615)} \\ \hline
\begin{tabular}[c]{@{}c@{}} Monolingual English\end{tabular} & 400, 000 & \multicolumn{2}{c|}{9,918,380 (266,640)} \\ \hline\hline
dev & 6, 970 & - & -\\ \hline
test & 6, 750 & - & - \\ \hline
\end{tabular}
\end{table}

\subsection{Data}
\label{sec:data}

We used the data from the IWSLT 2014 German-English shared translation task \cite{Cettolo2014} in this work. For pre-processing, we used the script provided by \cite{Ranzato2016} for data clean-up and splitting of train, dev and test sets. This results in 153,348; 6,970; and 6,750 parallel sentences for training, development and testing respectively. For the monolingual data, we randomly selected 400,000 English monolingual sentences of the pre-processed \cite{Luong2015} WMT 2014 English-German translation task \cite{Bojar2017}. Table~\ref{tab1} shows the statistics of the dataset. We learned byte pair encoding (BPE) \cite{Sennrich2016b} with 10,000 merge operations on the training dataset, applied it on the train, development and test datasets and, afterwards, we build the vocabulary of the training dataset. We train each of the self-learning and back-translation models using the \emph{"joint BPE"} approach as implemented in \cite{Kocmi2018}, mixing the authentic and synthetic data to learn a joint BPE and to build the training vocabulary, as shown to be superior in the works of \cite{Abdulmumin2021}, \cite{Abdulmumin2019a} and \cite{Kocmi2018}.

This work was implemented on only the German-English low resource translation dataset. We also use this data because the language pair have a quality estimation system, enabling the implementation of two out of the three proposed methods. This, therefore, enables a more comprehensive evaluation of the proposed methods against themselves and other current systems. The self-learning improved back-translation approach as preliminarily implemented in \cite{Abdulmumin2021} was shown to be successful at improving the performance of the backward model. It was also shown to generalize well on other low resource languages.

\subsection{Evaluation}
The work was evaluated using the BLEU metric. This is the most commonly used metric to determine the accurate a translation. Several works have found that the metric correlates with human translation \cite{Koehn2004, Babych2014, Chatzikoumi2020}. This means, therefore, that an improvement in the BLEU score of a translation system actually means that the system has improved. The metric works by counting and comparing n-grams that are present in the generated translations to the n-grams in the reference (test) set. Given the precision of n-grams of size up to N (the size is usually 1 to 4 grams as it has shown to generate the highest correlation with human evaluation \cite{Papineni2002}), \(p_n\), and the lengths of the test set, \(c\), and reference translations, \(r\), both in words,

\begin{equation}
	BLEU = b \cdot exp\left(\sum_{n=1}^{4}\log(p_n)\right)
\end{equation}

\begin{equation}
	b = min(1, e^{1-r/c}).
\end{equation}

In the computation of the BLEU score, a precision metric, the \emph{brevity penalty}, \(b\), is used instead of recall to penalize outputs that are too short compared to the reference translations. \(b\) ensures that the system does not translate only sections of the test set in which it is most confident and neglecting the other fragments, resulting in higher precision scores.

The models were evaluated on the development set after every 5,000 training steps (saved checkpoints). The training for each model is stopped when the maximum average improvement observed after evaluating four consecutive checkpoints is less than 0.2 BLEU. Unless stated otherwise, there was no extra training for any model after the stopping condition was met. After training, the checkpoints were evaluated on the test set to determine the best performing checkpoint on an unseen data. In most of the models trained, we obtained the best performance on the test set after averaging the last 8 checkpoints.

The work (proposed approaches) were evaluated against current systems in addition to the baselines. These current systems are: standard back-translation \cite{Sennrich2016a}, self-learning improved back-translation \cite{Abdulmumin2021} and the iterative back-translation \cite{Hoang2018}. The performance scores were also compared to other works that used the same training and test data.

\subsection{Models}

For comparing the performances of the backward models, we train the following models.

\renewcommand{\labelitemi}{\textbullet}
\begin{itemize}
	\item The \emph{baseline} backward model is trained on the available authentic parallel data without using the available monolingual data for improvement. It is considered the backward model in standard back-translation. In this work, we used the baseline in  \cite{Abdulmumin2021}.
	\item The \emph{backward\_ft} is the backward model trained on the available authentic parallel data and thereafter self-trained on all the generated synthetic data. The \emph{ft} indicates that the self-training approach was used to improve the model through forward translation.
	\item The \emph{backward\_ft + QE} is the backward model trained on the available authentic parallel data and thereafter self-trained on the best generated synthetic data selected using the QE system.
	\item The \emph{backward\_ibt} is the backward model improved using the iterative back-translation approach.
	\item The \emph{backward\_isl + QE} is the backward model improved using the iterative self-learning with Quality Estimation  approach as proposed in \cite{Specia2018}.
	\item The \emph{backward\_isl} is the backward model improved using the iterative self-learning approach without Quality Estimation.
	\item Finally, various forward models were trained, including the baseline forward model, the standard back-translation forward model and the corresponding forward models for all of the various backward model that were trained.
\end{itemize}

\section{Experiments and Results}
\label{sec:results}
In this section, we presented the experiments that were conducted to evaluate the effectiveness of the proposed hybrid approach with and without quality estimation and also to evaluate the ability of the approach to generalize on other low-resource languages. We also explained the experimental results obtained especially as we used the proposed approaches to improve the performance of the backward model. The results, in BLEU scores, of the various strategies on the backward model are shown in Tables~\ref{tab2} and \ref{tab3}. Finally, the results obtained on the forward models after using the various backward models to improve their performances were explained and are shown Tables~\ref{tab4} and \ref{tab5}.

\subsection{Baseline vs Self-Training with and without Quality Estimation}
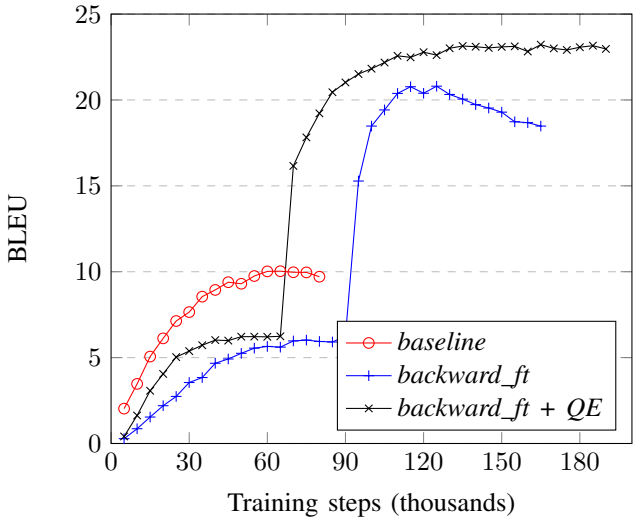
\begin{figure}[t!]
\centering
\begin{tikzpicture}
\begin{axis}[
    xlabel={Training steps (thousands)},
    ylabel={BLEU},
    xmin=0, xmax=200,
    ymin=0, ymax=25,
    xtick={0,30,60,90,120,150,180},
    ytick={0,5,10,15,20,25},
    legend pos=south east,
    legend cell align={left},
    ymajorgrids=true,
    grid style=dashed,
]
 
\addplot[
    color=red,
    mark=o
    ]
    table {baseline.txt};
    \addlegendentry{\emph{baseline}}

\addplot[
    color=blue,
    mark=+
    ]
    table {sl.txt};
    \addlegendentry{\emph{backward\_ft}}
    
\addplot[
    color=black,
    mark=x
    ]
    table {slqe.txt};
    \addlegendentry{\emph{backward\_ft + QE}}

\end{axis}
\end{tikzpicture}
\caption{Baseline vs Self-Training with and without Quality Estimation. Evaluation scores on the test set.}
\label{fig:2}
\end{figure}

This section presents the experiments and evaluations of the proposed approaches. We compare the performances of the baseline backward model with that of the backward models that were self-trained on the synthetic data with and without quality estimation. These performances are presented in Table~\ref{tab2} and Fig.~\ref{fig:2}.

\begin{table}[t]
\renewcommand{\arraystretch}{1.3}
\caption{Performances of the backward models in Self-Training approaches compared to the baseline. The scores marked * are placed under the fine-tune column for comparison only but the model was not fine-tuned as it was trained using the authentic data only.}
\label{tab2}
\begin{tabular}{|c||c||c||c||c|c|}
\hline
\multirow{2}{*}{models} & \multicolumn{2}{c||}{\begin{tabular}[c]{@{}c@{}}best models \\ (training steps) \end{tabular}} & \multicolumn{2}{c|}{\begin{tabular}[c]{@{}c@{}}average of last \\ 8 checkpoints \end{tabular}} \\ \cline{2-5}
 & pre-train & fine-tune & pre-train & fine-tune \\ \hline
\begin{tabular}[c]{@{}c@{}}\emph{baseline} \cite{Abdulmumin2021}\\ (Standard BT)\end{tabular} & - & \begin{tabular}[c]{@{}c@{}}10.03 \\ (65k)* \end{tabular} & - & 10.25* \\ \hline
\emph{backward\_ft} \cite{Abdulmumin2021} & \begin{tabular}[c]{@{}c@{}}6.02 \\ (75k)\end{tabular}  & \begin{tabular}[c]{@{}c@{}}20.77 \\ (115k)\end{tabular} & 6.01 & 21.31 \\ \hline
\begin{tabular}[c]{@{}c@{}}\emph{backward\_ft + QE}\end{tabular} & \begin{tabular}[c]{@{}c@{}}6.24 \\ (65k)\end{tabular} & \begin{tabular}[c]{@{}c@{}}23.22 \\ (165k)\end{tabular}& 6.23 & 23.66 \\ \hline
\end{tabular}
\end{table}

\begin{table*}
\renewcommand{\arraystretch}{1.3}
\centering
\caption{Performances of the backward models in the Iterative Back-Translation and various Iterative Self-Training approaches.}
\label{tab3}
\begin{tabular}{|c||c||c||c||c||c|}
\hline
\multirow{2}{*}{models} & \multirow{2}{*}{iterations} & \multicolumn{2}{c||}{\begin{tabular}[c]{@{}c@{}}best models \\ (training steps) \end{tabular}} & \multicolumn{2}{c|}{\begin{tabular}[c]{@{}c@{}}average of last \\ 8 checkpoints \end{tabular}} \\ \cline{3-6}
 & & pre-train & fine-tune & pre-train & fine-tune \\ \hline
\multirow{2}{*}{\emph{backward\_ibt}} & 1 & 2.64 (145k)  & 21.81 (200k) & 2.66 & 21.49 \\ \cline{2-6}
 & 2 & 7.88 (140k) & 23.72 (215k) & 7.94 & 23.89 \\ \hline\hline
\multirow{2}{*}{\begin{tabular}[c]{@{}c@{}}\emph{backward\_isl + QE} \end{tabular}} & 1 & 16.18 (105k) & 23.41 (200k) & 15.88 & 23.62 \\ \cline{2-6}
 & 2 & 19.59 (155k) & 23.69 (210k) & 19.83 & 24.05 \\ \hline\hline
\multirow{2}{*}{\begin{tabular}[c]{@{}c@{}}\emph{backward\_isl}\end{tabular}} & 1 & 18.57 (90k)  & 23.78 (190k) & 19.16 & 24.06 \\ \cline{2-6}
 & 2 & 21.52 (155k) & 24.03 (215k) & 21.51 & 24.27 \\ \hline
\end{tabular}
\end{table*}

\begin{figure*}
\centering
\begin{minipage}[t]{0.49\textwidth}
\resizebox{0.95\linewidth}{!}{%
\begin{tikzpicture}
\begin{axis}[
    xlabel={Training steps (thousands)},
    ylabel={BLEU},
    xmin=0, xmax=220,
    ymin=0, ymax=26,
    xtick={0,30,60,90,120,150,180, 210},
    ytick={0,5,10,15,20,25},
    legend pos=south east,
    legend cell align={left},
    ymajorgrids=true,
    grid style=dashed,
]
 
\addplot[
    color=black,
    mark=+
    ]
    table {ibtbm1.txt};
    \addlegendentry{\emph{bibt}}

\addplot[
    color=red,
    mark=-
    ]
    table {islqebm1.txt};
    \addlegendentry{\emph{bislqe}}
  
\addplot[
    color=green,
    mark=x
    ]
    table {islbm1.txt};
    \addlegendentry{\emph{bisl}}
    
%
%

\end{axis}
\end{tikzpicture}
}
\centering
~\textbf{a}. Iteration 1
\end{minipage}%
\hfill
\begin{minipage}[t]{0.49\textwidth}
\resizebox{0.95\linewidth}{!}{%
\raggedleft
\begin{tikzpicture}
\begin{axis}[
    xlabel={Training steps (thousands)},
    ylabel={BLEU},
    xmin=0, xmax=240,
    ymin=0, ymax=26,
    xtick={0,30,60,90,120,150,180, 210, 240},
    ytick={0,5,10,15,20,25},
    legend pos=south east,
    legend cell align={left},
    ymajorgrids=true,
    grid style=dashed,
]

\addplot[
    color=black,
    mark=+
    ]
    table {ibtbm2.txt};
    \addlegendentry{\emph{bibt}}    

\addplot[
    color=red,
    mark=-
    ]
    table {islqebm2.txt};
    \addlegendentry{\emph{bislqe}}

\addplot[
    color=green,
    mark=x
    ]
    table {islbm2.txt};
    \addlegendentry{\emph{bisl}}
 
%
%
   
\end{axis}
\end{tikzpicture}
}
\centering
~\textbf{b}. Iteration 2
\end{minipage}
\caption{Performances on the test set, after first and second iterations, of the different iterative approaches for improving the backward model. KEY: bibt = \emph{backward\_ibt}, bislqe = \emph{backward\_isl + QE} and bisl = \emph{backward\_isl}}
\label{fig3}
\end{figure*}
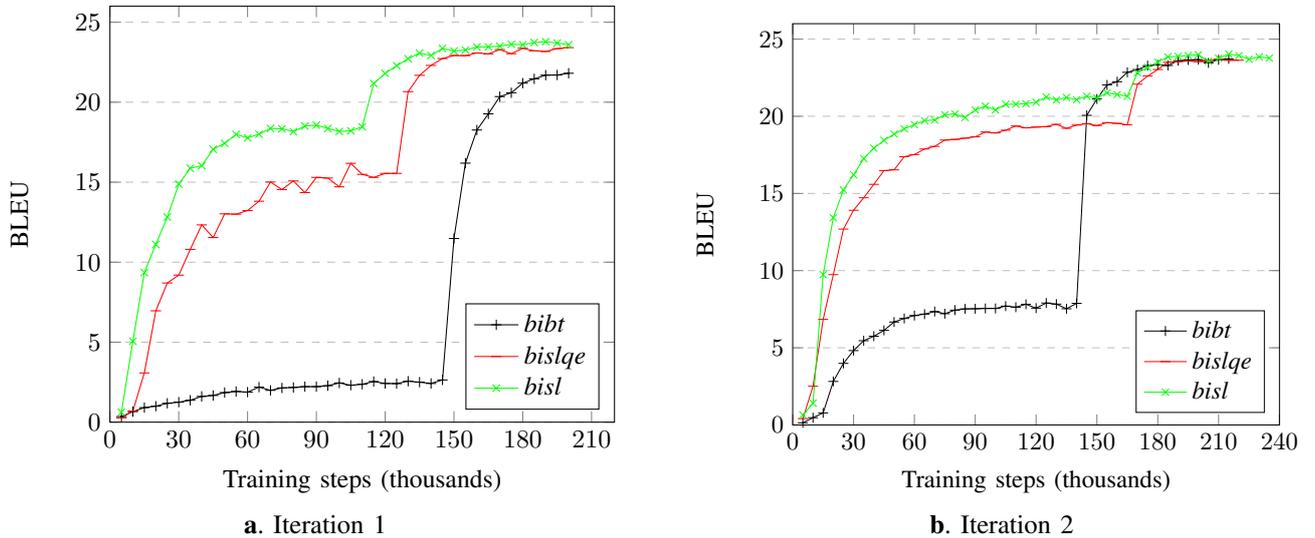

We used the backward model \cite{Abdulmumin2021}, En-De, that was trained on the available authentic parallel data. This model was trained for 80,000 training steps before the stopping condition was met and after evaluation, the checkpoint at the 65,000$^{th}$ step performed better than all the others, obtaining a BLEU score of 10.03. Having averaged the last 8 checkpoints, a better performance of 10.25 BLEU was achieved. The average model was used to generate the synthetic parallel data, translating the available monolingual target data. This synthetic parallel data was used in this work as the basis for all the experiments conducted. The two data -- synthetic and available authentic parallel data -- were used to train an improved backward model, \emph{backward\_ft}.

It can be observed that the performance of \emph{backward\_ft}, 20.77 BLEU, after stopping training at 165,000 steps, is more than double the performance of \emph{baseline} -- a 10.74 BLEU improvement over the best single checkpoint baseline model. The average of the last 8 checkpoints performed even better, obtaining a BLEU score of 21.31 (+0.54 BLEU). Finally, we used the Quality Estimation system described in Section~\ref{sec:setup} to select a subset of the synthetic sentences that are adjudged to be better translated than the others. We selected the best one-third of this data and together with the available authentic data, we trained another backward model, \emph{backward\_ft + QE}, using the pre-training and fine-tuning approach. This model achieved a $+2.35$ BLEU improvement over \emph{backward\_ft}.

It can be observed that although only a third of the synthetic sentences were added to the authentic training data, but since their quality was estimated to be better than the other two-thirds, the effect in reducing the quantity of the training data was not felt, they even yielded a better models. This huge gap in performance can be seen in Fig.~\ref{fig:2} where the self-learning approach with quality estimation performed better than the \emph{backward} and \emph{backward\_ft}, models that were trained on all of the authentic and authentic$+$synthetic data respectively. Even during pre-training, using the more qualitative subset was shown to train a better backward model that using all of the synthetic data. The performances of all the models were very poor, though, during pre-training but after after training on the authentic data -- fine-tuning -- they increased tremendously.

\subsection{The Iterative Approaches}
We then implemented the various iterative self-learning approaches that were aimed at improving the performance of the backward models. To evaluate the performance of our proposed iterative approaches over the current iteration in back-translation, we implemented the rather successful iterative back-translation of \cite{Hoang2018} using the synthetic data pre-training of \cite{Abdulmumin2019a}. The performances, measured in BLEU scores, of the backward models that were trained using this and the various iterative methods described in this work -- iterative self-learning with quality estimation \cite{Specia2018}, iterative self-learning without quality estimation (proposed in this work) -- are presented in Table~\ref{tab3} and plotted in Fig.~\ref{fig3}.

\begin{table*}
\renewcommand{\arraystretch}{1.3}
\centering
\caption{Performances of the forward models (De-En) trained using different quality of synthetic data compared to the baseline. The scores marked * are placed under the fine-tune column for comparison only but the model was not fine-tuned as it was trained using the authentic data only.}
\label{tab4}
\begin{tabular}{|c||c||c||c||c|c|}
\hline
\multirow{2}{*}{\begin{tabular}[c]{@{}c@{}}backward \\ models \end{tabular}} & \multicolumn{2}{c||}{\begin{tabular}[c]{@{}c@{}}best models \\ (training steps) \end{tabular}} & \multicolumn{2}{c|}{\begin{tabular}[c]{@{}c@{}}average of last \\ 8 checkpoints \end{tabular}} \\ \cline{2-5}
 & pre-train & fine-tune & pre-train & fine-tune \\ \hline
 none & -  & \begin{tabular}[c]{@{}c@{}}20.30 (75k)* \end{tabular}& - & 20.95* \\ \hline
\begin{tabular}[c]{@{}c@{}}\emph{baseline} \\ (Standard BT)\end{tabular} & \begin{tabular}[c]{@{}c@{}}5.43 (60k) \end{tabular}& \begin{tabular}[c]{@{}c@{}}28.31 (155k) \end{tabular}& 5.13 & 28.83 \\ \hline
\emph{backward\_ft} & \begin{tabular}[c]{@{}c@{}}17.60 (120k) \end{tabular}& \begin{tabular}[c]{@{}c@{}}29.45 (185k) \end{tabular}& 17.72 & 29.91 \\ \hline
\begin{tabular}[c]{@{}c@{}}\emph{backward\_ft + QE}\end{tabular} & \begin{tabular}[c]{@{}c@{}}19.38 (160k) \end{tabular}& \begin{tabular}[c]{@{}c@{}}30.01 (225k) \end{tabular}& 19.37 & 30.39 \\ \hline
\end{tabular}
\end{table*}

\begin{table*}
\renewcommand{\arraystretch}{1.3}
\centering
\caption{Performances of Forward Models Trained using Iterative Back-Translation and Iterative Self-Training Enhanced Back-Translation approaches.}
\label{tab5}
\begin{tabular}{|c||c||c||c||c||c|}
\hline
\multirow{2}{*}{\begin{tabular}[c]{@{}c@{}}backward \\ models \end{tabular}} & \multirow{2}{*}{iterations} & \multicolumn{2}{c||}{\begin{tabular}[c]{@{}c@{}}best models \\ (training steps) \end{tabular}} & \multicolumn{2}{c|}{\begin{tabular}[c]{@{}c@{}}average of last \\ 8 checkpoints \end{tabular}} \\ \cline{3-6}
 & & pre-train & fine-tune & pre-train & fine-tune \\ \hline
\multirow{2}{*}{\emph{backward\_ibt}} 
 & 1 & 17.87 (160k)  & 29.62 (215k) & 18.04 & 30.01 \\ \cline{2-6}
 & 2 & 18.82 (135k) & 29.71 (215k) & 18.94 & 30.14 \\ \hline\hline
\multirow{2}{*}{\begin{tabular}[c]{@{}c@{}}\emph{backward\_isl + QE}\end{tabular}}
 & 1 & 15.90 (135k) & 29.22 (180k) & 16.18 & 29.62 \\ \cline{2-6}
 & 2 & 17.97 (120k) & 29.55 (185k) & 18.09 & 30.05 \\ \hline\hline
\multirow{2}{*}{\begin{tabular}[c]{@{}c@{}}\emph{backward\_isl}\end{tabular}}
 & 1 & 18.92 (130k) & 30.00 (180k) & 19.09 & 30.27 \\ \cline{2-6}
 & 2 & 19.21 (140k) & 29.88 (185k) & 19.26 & 30.33 \\ \hline
\end{tabular}
\end{table*}

We first implemented the iterative back-translation approach to determine the performance of the backward model -- \emph{backward\_ibt} -- after two iterations (the number of iterations compared in this work). The first iteration produced a good model, though under-performing the self-iteration with quality estimation by $-2.17$ BLEU. But after the second iteration, the model caught up and even outperformed the former by $+0.23$ BLEU. The difference in performance was not very significant compared to: \textbf{(i)} the number of models that were required to be trained -- 2 backward models to obtain the performance reached by \emph{backward\_ft + QE} vs 5 (3 backward and 2 forward) models to obtain the performance reached by \emph{backward\_ibt}; and \textbf{(ii)} the time taken to train these models (\emph{backward\_ft + QE} trained for a cumulative of $0.27M$ steps while \emph{backward\_ibt} trained for $0.87M$) and to generate the synthetic data at each stage of the back-translation processes.

We then implemented the iterative self-learning approach with quality estimation as described in Section~\ref{sec:islqe} to train another backward model: \emph{backward\_isl + QE}. After two iterations, the approach gained a +0.16 BLEU over \emph{backward\_ibt} after training a cumulative four backward models for $0.69M$ training steps. Finally, the iterative self-learning approach without quality estimation achieved the best score, outperforming \emph{backward\_isl + QE} by +0.22 BLEU. The best performing backward model was, therefore, trained using the iterative self-learning approach without quality estimation and was obtained faster (requiring the same number of models but trained cumulatively for $0.68M$ training steps), requiring the training of less number of models than \emph{backward\_ibt} and eliminating the need for the time-consuming quality estimation of \emph{backward\_isl + QE} -- an estimated time of over 2 days.

In our proposed approaches, the quality of the synthetic data continues to improve after each iteration -- see BLEU scores from Table~\ref{tab3}. Though the self-training without quality estimation underperformed the self-training with quality estimation at iteration 0 -- see Table~\ref{tab2} -- subsequent self-training iterations resulted in rapid improvement in the performance of the backward model -- see Table~\ref{tab3}. It can also be seen that at the second (final) iteration, the \emph{backward\_isl} model pre-trained on only the synthetic data outperformed the \emph{backward\_ft} model that was trained on both the authentic and synthetic parallel data by +0.2 BLEU score (21.51 BLEU vs 21.31 BLEU), and the \emph{baseline} that was trained on the authentic data by +11.26 BLEU score. This has shown, therefore, that the quality of synthetic data can reach a point where it can train a better model than a low resourced model or even low resourced but self-trained model.

\subsection{Back-Translation}
\label{bt}

Tables~\ref{tab4} and \ref{tab5} show the performance evaluation in BLEU scores of the forward models trained with and without synthetic data. All the models that are trained with synthetic data outrightly outperformed the baseline NMT. For the backward models trained with synthetic data, using back-translation, our approaches achieved significant improvements over the works we compared: standard back-translation \cite{Sennrich2016}, self-training enhanced back-translation of \cite{Abdulmumin2021} and the iterative back-translation of \cite{Hoang2018}. But the differences in the performances between the three backward models that were trained using the proposed methods were not very large and, therefore, the corresponding forward models were also expected to achieve similar performances.

To compare the quality of the forward models, we first trained a standard NMT forward model on the available parallel data only. This model achieved the best performance of 20.95 BLEU after averaging the last 8 checkpoints based on the set-up and the stopping criteria described in Section~\ref{sec:setup}. As mentioned previously, the back-translation benefits most when the quality of the synthetic data is close to that of the authentic parallel data as shown in studies such as \cite{Edunov2018,Poncelas2018,Cotterell2018}. The possibility of getting a better synthetic data lies in training a good backward model. To test this claim, we train various forward models on the authentic parallel data and the different synthetic data that were generated using the various backward models trained in this work. We used all of the generated synthetic data to train the forward models unlike in the training of some of the backward models where only some of the data were selected and used for self-learning. Experimental results in Tables~\ref{tab4} and \ref{tab5} showed that the gain observed in the performances of the self-trained models manifested ultimately in the qualities of the forward models.

We observed improvements of 9.44, 1.56, 0.48; 9.1, 1.22, 0.14; and 9.38, 1.5, 0.42 BLEUs in the proposed methods -- the self-training with quality estimation, iterative self-training with quality estimation and iterative self-training -- over the baseline NMT, standard back-translation and self-trained back-translation respectively. Between the proposed approaches, however, the forward model in the self-training with quality approach performed better than the other proposed methods, achieving 0.34 and 0.06 BLEUs improvement over the iterative self-learning with and without quality estimation respectively. The approach also performed better than the iterative back-translation by 0.25 BLEU.

Finally, Table~\ref{tab10} shows how our work compares to current works that uses the same training and test data. Using the same architecture, the following works trained translation models with varying degree of performance. Ranzato et al. \cite{Ranzato2016} trained the first benchmark on the dataset. \cite{Wiseman2016} introduced beam-search optimization to improve on the benchmark. \cite{Bahdanau2017} proposed a reinforcement learning ``actor-critic'' methodology where the critic helps the actor -- the main sequence prediction network. Finally, \cite{Huang2018} hybridized the neural and phrase-based architectures to improve the performance on the data. Our works compares well with these works, with the hybrid of self-learning with quality estimation and back-translation achieving a better performance than the next best by $+0.31$. On the other hand, \cite{Shen2020} used the more advanced transformer architecture to train the current state-of-the-art translation model on this data. The authors proposed a data augmentation strategy called ``cutoff'', where a section of the information in the input sentence is deleted to give its ``restricted'' views when fine-tuning.

\section{Discussion}
\label{sec:discuss}

Neural machine translation systems relies on a huge amount of parallel data to train standard, state-of-the-art translation models. For low resource languages, these models perform woefully when deployed. Back-translation is an approach that was introduced in NMT by \cite{Sennrich2016} to enable the generation of additional data for improving translation in both low and high resource languages. Subsequent studies have shown that the approach require other special methods to reach an acceptable standard for translation quality especially in low resource set-ups \cite{Hoang2018,Dabre2019,Kocmi2019,Niu2018}. In these set-ups, the backward model is trained on a scarce data and, therefore, the quality of generated additional data may not be enough to substantially improve the target translation model. The target of back-translation is always to improve the performance of the forward model on the available monolingual data and not the intermediary backward model. But the standard of the forward model relies on the authentic data and the ability of the backward model to generate a good enough additional training data.

\begin{table}
\renewcommand{\arraystretch}{1.3}
\centering
\caption{Available systems trained on the IWSLT'14 German-English data.}
\label{tab10}

\begin{tabular}{|c|c|c|c|}
\hline
System & Architecture & Method & Performance \\ \hline

\cite{Ranzato2016} & RNMT & MIXER & 21.83 \\ \hline

\cite{Wiseman2016} & RNMT & BSO & 25.48 \\ \hline

\cite{Bahdanau2017} & RNMT & Actor-Critic & 28.53 \\ \hline

\cite{Huang2018} & Neural PBMT & Neural PBMT + LM & 30.08 \\ \hline \hline

\multirow{3}{*}{\textbf{\emph{This work}}} & \multirow{3}{*}{RNMT} & ISL+QE+BT & 30.05 \\ \cline{3-4}

 & & ISL+BT & 30.33 \\ \cline{3-4}
 
 & & SL+QE+BT & \textbf{30.39} \\ \hline \hline
 
\cite{Shen2020} & Transformer & Cutoff & \textbf{37.60} \\ \hline

\end{tabular}
\end{table}

\begin{table*}
\renewcommand{\arraystretch}{1.3}
\centering
\caption{This table shows how often a conclusion with 95\% statistical significance is made for comparing the various backward models. We used different sample sizes of 100, 500 and 1000 sentences for each of the approach on English-German low resource NMT.}
\label{tab6}
\begin{tabular}{|l||c||c||c||c||c|}
  \hline
  \multirow{2}{*}{System Comparison} & \multirow{2}{*}{\begin{tabular}[c]{@{}c@{}}BLEU \\ difference\end{tabular}} & \multicolumn{4}{c|}{Sample size} \\ \cline{3-6}
  & & 50 & 100 & 500 & 1000 \\ \hline
  \emph{backward\_ft} is better than \emph{baseline} & 11.06 &100\%& 100\%     & 100\%    & 100\%  \\ \hline
  \emph{backward\_ft + QE} is better than \emph{baseline} & 13.41 &100\%& 100\%     & 100\%    & 100\% \\ \hline
  \emph{backward\_ibt} is better than \emph{baseline} & 13.64 &100\%& 100\%     & 100\%    & 100\% \\ \hline
  \emph{backward\_isl + QE} is better than \emph{baseline} & 13.80 &100\%& 100\%     & 100\%    & 100\% \\ \hline
  \emph{backward\_isl} better than \emph{baseline} & 14.02 &100\%& 100\%     & 100\%    & 100\% \\ \hline
  \emph{backward\_ft + QE} is better than \emph{backward\_ft} & 2.35 &100\%& 100\%     & 100\%    & 100\% \\ \hline
  \emph{backward\_ibt} is better than \emph{backward\_ft} & 2.58 &100\%& 100\%     & 100\%    & 100\% \\ \hline
  \emph{backward\_isl + QE} is better than \emph{backward\_ft} & 2.74 &100\%& 100\%     & 100\%    & 100\% \\ \hline
  \emph{backward\_isl} better than \emph{backward\_ft} & 2.96 &100\%& 100\%     & 100\%    & 100\% \\ \hline
  \emph{backward\_ibt} is better than \emph{backward\_ft + QE} & 0.23 &94\%& 94\%      & 96\%     & 94.4\% \\ \hline
  \emph{backward\_isl + QE} is better than \emph{backward\_ft + QE} & 0.39 &100\%& 100\%     & 100\%    & 99.8\% \\ \hline
  \emph{backward\_isl} better than \emph{backward\_ft + QE} & 0.61 &100\%& 100\%    & 100\%   & 100\% \\ \hline
  \emph{backward\_isl + QE} is better than \emph{backward\_ibt} & 0.16 &86\%& 89\%     & 88.8\%    & 87.4\% \\ \hline
  \emph{backward\_isl} better than \emph{backward\_ibt} & 0.38&100\%& 100\%    & 99.6\%   & 99.7\% \\ \hline
  \emph{backward\_isl} better than \emph{backward\_isl + QE} & 0.22 &94\%& 92\%    & 93.4\%   & 94.9\% \\ \hline
     \end{tabular}
\end{table*}

\begin{table*}
\renewcommand{\arraystretch}{1.3}
\centering
\caption{This table shows how often a conclusion with 95\% statistical significance is made for comparing the various forward models that were trained with and without the various synthetic data. We used different sample sizes of 100, 500 and 1000 sentences for each of the approach on English-German low resource NMT. KEY: SLQEBT = self-learning with quality estimation enhanced back-translation}
\label{tab7}
\begin{tabular}{|p{10.5cm}||c||c||c||c||c|}
  \hline
  \multirow{2}{*}{System Comparison} & \multirow{2}{*}{\begin{tabular}[c]{@{}c@{}}BLEU \\ difference\end{tabular}} & \multicolumn{4}{c|}{Sample size} \\ \cline{3-6}
  & & 50 & 100 & 500 & 1000 \\ \hline
  \emph{standard back-translation} is better than \emph{baseline} & 7.89 &100\%& 100\%     & 100\%    & 100\%  \\ \hline
  \emph{self-learning enhanced back-translation} is better than \emph{baseline} & 8.96 &100\% &100\%     & 100\%    & 100\% \\ \hline
  \emph{SLQEBT} is better than \emph{baseline} & 9.44 &100\%& 100\%      & 100\%     & 100\% \\ \hline
  \emph{iterative back-translation} is better than \emph{baseline} & 9.19 &100\%& 100\%     & 100\%    & 100\% \\ \hline
  \emph{iterative SLQEBT} better than \emph{baseline} & 9.10 &100\%& 100\%    & 100\%   & 100\% \\ \hline
  \emph{iterative self-learning enhanced back-translation} better than \emph{baseline} & 9.38 &100\%& 100\%    & 100\%   & 100\% \\ \hline
  
  \emph{self-learning enhanced back-translation} is better than \emph{standard back-translation} & 1.08 &100\%& 100\%     & 100\%    & 100\% \\ \hline
  \emph{SLQEBT} is better than \emph{standard back-translation} & 1.56 &100\%& 100\%      & 100\%     & 100\% \\ \hline
  \emph{iterative back-translation} is better than \emph{standard back-translation} & 1.31 &100\%& 100\%     & 100\%    & 100\% \\ \hline
  \emph{iterative SLQEBT} better than \emph{standard back-translation} & 1.22 &100\%& 100\%    & 100\%   & 100\% \\ \hline
  \emph{iterative self-learning enhanced back-translation} better than \emph{standard back-translation} & 1.49 &100\%& 100\%    & 100\%   & 100\% \\ \hline
  
  \emph{SLQEBT} is better than \emph{self-learning enhanced back-translation} & 0.48 &100\%& 100\%      & 100\%     & 100\% \\ \hline
  \emph{iterative back-translation} is better than \emph{self-learning enhanced back-translation} & 0.23 &96\%& 97\%     & 95.6\%    & 95.7\% \\ \hline
  \emph{iterative SLQEBT} better than \emph{self-learning enhanced back-translation} & 0.14 &86\%& 86\%    & 85\%   & 86.4\% \\ \hline
  \emph{iterative self-learning enhanced back-translation} better than \emph{self-learning enhanced back-translation} & 0.42 &100\%& 100\%    & 100\%   & 100\% \\ \hline
  
  \emph{iterative back-translation} better than \emph{iterative SLQEBT} & 0.09 &78\%& 77\%    & 74\%   & 74.9\% \\ \hline
  \emph{iterative self-learning enhanced back-translation} better than \emph{iterative SLQEBT} & 0.28 &98\%& 99\%    & 98.2\%   & 99.1\% \\ \hline
  \emph{SLQEBT} better than \emph{iterative SLQEBT} & 0.34 &100\%& 100\% & 99.2\% & 99.9\% \\ \hline
  
  \emph{iterative self-learning enhanced back-translation} better than \emph{iterative back-translation} & 0.19 &96\%& 92\%    & 91\%   & 93.5\% \\ \hline
  \emph{SLQEBT} is better than \emph{iterative back-translation} & 0.25 &100\%& 98\% & 96.2\%    & 97.6\% \\ \hline
  
  \emph{SLQEBT} better than \emph{iterative self-learning enhanced back-translation} & 0.06 &60\%& 69\% & 66\% & 67.8\% \\ \hline
     \end{tabular}
\end{table*}

This work, therefore, presented a new variant of the back-translation that incorporates the self-learning approach, through forward translation, to use the same target-side monolingual data to improve not only the forward model, but the backward model also. The back-translation was used ultimately to improve the forward model but only after using self-training to enhance the standard of the backward model.

In implementing the self-learning approach, we investigated various methods namely: self-training and iterative self-training each with and without quality estimation. The iterative self-learning without quality estimation enhanced back-translation approach was proposed to avoid being so much reliant on the availability or reliability thereof of the quality estimation systems for the successful implementation of the previously proposed approaches. We determined that without such systems readily available, retraining the backward model over some iterations on all of the available synthetic data is capable of achieving the same or even superior performance to the other quality estimation methods. We implemented all the methods using the pre-training and fine-tuning strategies of \cite{Abdulmumin2019a} to enable each model differentiate between synthetic and authentic data during training, as this has been shown to improve the performance of models trained in such settings \cite{Edunov2018,Sennrich2016,Caswell2019,Abdulmumin2019a}. All performance scores obtained through experiments have been shown to be statistically significant using the paired bootstrap resampling of \cite{koehn-2004-statistical} as implemented in \cite{NEURIPS2019_9015} -- see Tables~\ref{tab6} and \ref{tab7}.

\begin{table*}
\renewcommand{\arraystretch}{1.3}
\centering
\caption{An English-to-German backward model's translation example in the IWSLT-EnDe 14 test set.}
\label{tab8}
\begin{tabular}{|p{2.5cm}||p{10cm}|}
\hline
Source & right now we spend three billion hours a week playing online games \\ \hline

Reference & derzeit verbringen wir 3 milliarden stunden pro woche mit online-spielen \\ \hline\hline

\emph{baseline} & momentan \textbf{verbringen wir drei milliarden stunden pro woche} spiele in spiele spielen \\ \hline

\emph{backward\_ft} & jetzt \textbf{verbringen wir drei milliarden stunden pro woche mit online-spielern} \\ \hline

\emph{backward\_ft + QE} & jetzt \textbf{verbringen wir drei milliarden stunden pro woche mit} spielen spielen \\ \hline

\emph{backward\_ibt} & jetzt \textbf{verbringen wir drei milliarden stunden pro woche} , um online spiele zu spielen. \\ \hline

\emph{backward\_isl + QE} & jetzt \textbf{verbringen wir drei milliarden stunden pro woche online} spielspiele. \\ \hline

\emph{backward\_isl} & jetzt \textbf{verbringen wir drei milliarden stunden pro woche} für online-spiele. \\ \hline

\end{tabular}
\end{table*}

The work was evaluated on the low resource IWSLT'14 English-German neural machine translation. We observed that even though the proposed self-trained backward method \cite{Abdulmumin2021} outperformed the standard back-translation's backward model without using any of quality estimation or freezing of parameters in the decoder as proposed in \cite{Specia2018} and \cite{Zhang2016}, selecting and using the best synthetic data for self-training further improves its performance. This shows that although an improved performance was achieved, the full potential of the proposed method may not be realized when using vanilla self-training because the noise in the synthetic data will degrade the decoder's performance.

We extended the positive results that were obtained using self-learning by determining the benefit or otherwise of selecting only a fraction of the synthetic data for self-training using a quality estimation system. Experimental results indicated that not only was the result not affected by the reduction in training data, but that the performance was improved significantly, achieving $+2.35$ BLEU. We showed that not all of the synthetic data is required -- quantity -- but that the more beneficial additional data -- quality -- is just enough to train a superior backward model. Also, when the backward model -- or any other model -- is able to differentiate between the synthetic and authentic parallel data during training, then the effects of the lack of quality in the synthetic data becomes less problematic but the more the qualitative the synthetic data is, the better the model trained.

We also implemented an iterative approach that continued to enhance the quality of the backward model on the synthetic data. Each improved backward model was used to generate a synthetic training data for training the next improved model. The approach achieved a significant $+2.96$ BLEU improvement over the one-time usage of the self-training on the IWSLT'14 En-De test set. We compared our proposed iterative self-learning approach to other iterative approaches in \cite{Specia2018} and \cite{Hoang2018} and our method was shown to be superior while also requiring less number of models -- \((k-1)\)-less number of models in any \(k\textgreater0\) iterations -- to be trained than that needed in the approach of \cite{Hoang2018}. Also, unlike in \cite{Specia2018}, we showed that without data selection through the often time-consuming quality estimation, we achieved a similar performance.

Although the difference in quality observed in the models trained using the proposed approaches is not very significant -- with the best iterative approach out-performing the next best method by just over 0.2 BLEU, we noted that for the two iterations, the iterative back-translation required the training of 6 models (3 forward and 3 backward models) compared to the 5 models required in each of our iterative approaches. Also, while the iterative self-training in \cite{Specia2018} requires the same number of models to be trained as our proposed self-learning approach, a huge amount of time was taken to estimate the quality of the synthetic data. The cumulative time taken to estimate the quality of the synthetic data generated in the approach of \cite{Specia2018} was estimated at 2 days 6 hours (the data in iteration 1 took 31 hours, 1 minute and 29 seconds and; that in iteration 2 took 22 hours, 54 minutes and 53 seconds exactly).

It can be observed that among our proposed approaches, the best outperformed the previous works we compared in this work, namely: baseline backward model of standard back-translation by $+14.02$ BLEU. Also, each of the approach described in this work -- self-training with quality estimation, iterative self-training with quality estimation and iterative self-training enhanced back-translations -- achieved a better performance than the successful self-training enhanced back-translation of \cite{Abdulmumin2021} by $+2.35$, $+2.74$ and $+2.96$ BLEUs respectively. The iterative approaches we used in enhancing the quality of the backward model -- iterative self-training with quality estimation and iterative self-training enhanced back-translations -- outperformed the rather successful iterative back-translation by $+0.16$ and $+0.38$ BLEUs respectively.

\begin{table*}
\renewcommand{\arraystretch}{1.3}
\centering
\caption{A German-to-English forward model's translation example in the IWSLT-EnDe 14 test set.}
\label{tab9}
\begin{tabular}{|p{4.5cm}||p{12cm}|}
\hline
\begin{tabular}[c]{@{}c@{}} Source \end{tabular} & und es funktionierte . wieder hatten wir etwas magisches geschaffen . und die wirkung im publikum war dieselbe . allerdings haben wir mit dem film schon ein bisschen mehr geld eingespielt \\ \hline

Reference & and it did , and we created magic again , and we had the same result with an audience -- although we did make a little more money on that one. \\ \hline\hline

Baseline & and it worked . again . we had something magical , and the effect in the audience was the same thing , but we had a little more compound with a little more money. \\ \hline

Standard BT & and it worked . again , we did something like a magical thing , and the effect in the audience was the same thing , but we've done a little bit more money with the movie. \\ \hline

Self-learning enhanced BT & and it worked . again , we created something magical , and the effect in the audience was the same . but we had a little bit more money in the movie. \\ \hline

Self-Learning + QE enhanced BT & and it worked . again , we created something magical , and the effect in the audience was the same . but we had a little bit more money in the movie. \\ \hline

Iterative BT & and it worked . again , we created something magical , and the effect in the audience was the same . but we did the film a little bit more money. \\ \hline

Iterative Self-Learning + QE enhanced BT & and it worked . again , we had created something magical , and the effect in the audience was the same . but we played a little bit more money with the film. \\ \hline

Iterative Self-Learning enhanced BT & and it worked . again , we created something magical , and the effect in the audience was the same thing , but we had a little bit more money with the film. \\ \hline

\end{tabular}
\end{table*}

During the training of the forward models, while \cite{Edunov2018} suggests that models trained on synthetic data only can reach a performance similar that of models trained on the authentic data only, we showed that a model trained on a sufficient number of qualitative synthetic sentences can achieve a better performance than that of a model trained on low resource authentic parallel data. \cite{Fadaee2018} claimed that the ratio of synthetic to parallel data affects the translation model more than the quality of the backward model. This, they claimed, is because the model then tends to learn more from the synthetic data which often contain more noise. Instead, we claim that the quality of the backward model affects the performance of the approach more than the ratio because when the model is able to generate synthetic data that is close to or the same quality as human translation, the ratio of synthetic data to authentic data does not matter because the two data become more inseparable.

The forward models' performances were shown to reflect the improvements in the backward models. We achieved an improved $+0.48$ BLEU over the performance of self-trained enhanced back-translation method after using the QE system. All of the proposed approaches achieved better performances than all the previous methods but the quality between these approaches was observed to be similar as hypothesized in Section~\ref{bt}. This is as expected because the performances of the backward models were not far off from each other.

In Table~\ref{tab8}, we showed a sample translation from English to German. Our proposed models were able to produce exact translations to most  of the referenced translation: ''... wir 3 milliarden stunden pro woche mit online-spielen'' and the other part where the translation generated was different, the meaning was the same: ''derzeit'' 'vs' ''jetzt''. The self-trained models were able to generate exact translation to most of the referenced text but could only specify the adverb ''now'' instead of the referenced ''right now''. For the forward model, the effects of the improved backward models were observed in their performances. In Table~\ref{tab9}, we also translated a given German source text to English. The performances of the last two models (trained on the synthetic sentences generated by the backward model improved using our approach), and especially the last model, seemed to be more superior than the rest. The pre-training and fine-tuning approach has shown to be the better approach when applying the method we proposed in this work. As proposed in \cite{Abdulmumin2019a}, we found that pre-training first on the synthetic data and thereafter fine-tuning the model on the authentic data is the best strategy.

\section{Conclusion \& Future work}
\label{sec:conclude}

In this work, a hybrid approach for training better models in low resource neural machine translation is proposed. This category of languages have been shown to straggle their high resource counterparts even when the same methods are applied to improve their quality. The monolingual target data was used through self-learning and back-translation to complement the lack of enough training data. The back-translation approach that has shown tremendous potential for improving translation performance in high resource languages, has shown improved but less than desirable performance in low resource languages. This has been been as a result of the lack in quality of the backward model. The main essence of the hybridization is to efficiently utilize the advantage of the back-translation approach to generate the additional training data and that of the self-learning to ensure that the additional data is as close to the quality of the authentic data as possible.

A set of different implementations of the self-learning approach were investigated to determine not only the more effective, but also the easiest (feasible) method to apply on most low resource languages. The one-time self-learning with quality estimation and its iterative version as proposed in \cite{Ueffing2006} and \cite{Specia2018} respectively were implemented and compared to the versions that do not need estimating the synthetic data quality. These novel implementations were proposed to enable the implementation of the approach on languages where the quality estimation model or the data to train such systems is not available. Experimental results obtained on low resource English-German have shown that the proposed approach (and all of the various implementations) was able to train an improved backward model and the resulting synthetic training data was able to help in the training of superior target models than the widely successful back-translation approach.

The approach was shown to be straightforward and easy to implement on any low resource language translation to train a better model capable of attaining a more acceptable standard of translation. The self-training approach was shown to perform better when quality estimation is used to extract the best translations and used to retrain the generating backward model. For languages that do not have quality estimation systems, the iterative self-learning implementation has shown to be capable of achieving similar performances. The simplified iterative self-learning back-translation approach was shown to reduce both the number of models required, the time taken to achieve the number of iterations and the requirement of target and source monolingual data in iterative back-translation. We showed that it is possible to rely only on large amounts of synthetic data that gets improved iteratively especially in low resource conditions than strictly relying on the quality of fewer training data. We showed that the approach works well on a low resource neural machine translation.

For future work, we aim to determine the appropriate sentences to be considered fit for self-learning for each iteration especially using data selection as an alternative to quality estimation. We also intend to apply the approach on high resource languages.

\ifCLASSOPTIONcaptionsoff
  \newpage
\fi



%



\bibliographystyle{BibTeXtran}   
\bibliography{citations}       

\begin{thebibliography}{10}
\providecommand{\url}[1]{#1}
\csname url@samestyle\endcsname
\providecommand{\newblock}{\relax}
\providecommand{\bibinfo}[2]{#2}
\providecommand{\BIBentrySTDinterwordspacing}{\spaceskip=0pt\relax}
\providecommand{\BIBentryALTinterwordstretchfactor}{4}
\providecommand{\BIBentryALTinterwordspacing}{\spaceskip=\fontdimen2\font plus
\BIBentryALTinterwordstretchfactor\fontdimen3\font minus
  \fontdimen4\font\relax}
\providecommand{\BIBforeignlanguage}[2]{{%
\expandafter\ifx\csname l@#1\endcsname\relax
\typeout{** WARNING: IEEEtran.bst: No hyphenation pattern has been}%
\typeout{** loaded for the language `#1'. Using the pattern for}%
\typeout{** the default language instead.}%
\else
\language=\csname l@#1\endcsname
\fi
#2}}
\providecommand{\BIBdecl}{\relax}
\BIBdecl

\bibitem{Bahdanau2014}
D.~Bahdanau, K.~Cho, and Y.~Bengio, ``{Neural Machine Translation by Jointly
  Learning to Align and Translate},'' in \emph{3rd International Conference on
  Learning Representations, {\{}ICLR{\}} 2015, San Diego, CA, USA, May 7-9,
  2015, Conference Track Proceedings}, Y.~Bengio and Y.~LeCun, Eds., 2015.

\bibitem{Gehring2017}
J.~Gehring, A.~Michael, D.~Grangier, D.~Yarats, and Y.~N. Dauphin,
  ``{Convolutional Sequence to Sequence Learning},'' in \emph{Proceedings of
  the 34th International Conference on Machine Learning}, D.~Precup and Y.~W.
  Teh, Eds., Sydney, Australia, pp. 1243--1252, 2017.

\bibitem{Vaswani2017}
A.~Vaswani, N.~Shazeer, N.~Parmar, J.~Uszkoreit, L.~Jones, A.~N. Gomez,
  L.~Kaiser, and I.~Polosukhin, ``{Attention Is All You Need},'' in \emph{31st
  Conference on Neural Information Processing Systems}, Long Beach, CA, USA, pp. 6000--6010,
  2017.

\bibitem{Edunov2018}
S.~Edunov, M.~Ott, M.~Auli, and D.~Grangier, ``{Understanding Back-Translation
  at Scale},'' in \emph{Proceedings of the 2018 Conference on Empirical Methods
  in Natural Language Processing}.\hskip 1em plus 0.5em minus 0.4em\relax
  Brussels, Belgium: Association for Computational Linguistics, pp.
  489--500, 2018.

\bibitem{Hassan2018}
H.~Hassan, A.~Aue, C.~Chen, V.~Chowdhary, J.~Clark, C.~Federmann, X.~Huang,
  M.~Junczys-Dowmunt, W.~Lewis, M.~Li, S.~Liu, T.-y. Liu, R.~Luo, A.~Menezes,
  T.~Qin, F.~Seide, X.~Tan, F.~Tian, L.~Wu, S.~Wu, Y.~Xia, D.~Zhang, Z.~Zhang,
  and M.~Zhou, ``{Achieving Human Parity on Automatic Chinese to English News
  Translation},'' \emph{arXiv:1803.05567v3 [cs.CL]}, 2018.

\bibitem{Hoang2018}
V.~C.~D. Hoang, P.~Koehn, G.~Haffari, and T.~Cohn, ``{Iterative
  Back-Translation for Neural Machine Translation},'' in \emph{Proceedings of
  the 2nd Workshop on Neural Machine Translation and Generation}.\hskip 1em
  plus 0.5em minus 0.4em\relax Melbourne, Australia: Association for
  Computational Linguistics, pp. 18--24, 2018.

\bibitem{Ott2018}
M.~Ott, S.~Edunov, D.~Grangier, and M.~Auli, ``{Scaling Neural Machine
  Translation},'' in \emph{Proceedings of the Third Conference on Machine
  Translation: Research Papers}.\hskip 1em plus 0.5em minus 0.4em\relax
  Brussels, Belgium: Association for Computational Linguistics, pp. 1--9, 2018.

\bibitem{Abdalla2017}
M.~Abdalla and G.~Hirst, ``{Cross-lingual sentiment analysiswithout (good)
  translation},'' in \emph{Proceedings of the Eighth International Joint
  Conference on Natural Language Processing (Volume 1: Long Papers)}.\hskip 1em
  plus 0.5em minus 0.4em\relax Taipei, Taiwan: Asian Federation of Natural
  Language Processing, pp. 506--515, 2017.

\bibitem{Bracewell2008}
D.~B. Bracewell, F.~Ren, and S.~Kuroiwa, ``{A Low Cost Machine Translation
  Method for Cross-Lingual Information Retrieval},'' \emph{Engineering
  Letters}, vol.~16, no.~1, pp. 160--165, 2008.

\bibitem{Zoph2016}
B.~Zoph, D.~Yuret, J.~May, and K.~Knight, ``{Transfer Learning for Low-Resource
  Neural Machine Translation},'' in \emph{Proceedings of the 2016 Conference on
  Empirical Methods in Natural Language Processing}.\hskip 1em plus 0.5em minus
  0.4em\relax Austin, Texas: Association for Computational Linguistics,
  pp. 1568--1575, 2016.

\bibitem{Specia2018}
L.~Specia and K.~Shah, ``{Machine Translation Quality Estimation: Applications
  and Future Perspectives},'' in \emph{Translation Quality Assessment: From
  Principles to Practice}, J.~Moorkens, S.~Castilho, F.~Gaspari, and
  S.~Doherty, Eds.\hskip 1em plus 0.5em minus 0.4em\relax Cham: Springer
  International Publishing, pp. 201--235, 2018.

\bibitem{Ueffing2006}
N.~Ueffing, ``{Using Monolingual Source-Language Data to Improve MT
  Performance},'' in \emph{International Workshop on Spoken Language
  Translation}, Kyoto, Japan, pp. 174--181, 2006.

\bibitem{Zhang2016}
J.~Zhang and C.~Zong, ``{Exploiting Source-side Monolingual Data in Neural
  Machine Translation},'' in \emph{Proceedings of the 2016 Conference on
  Empirical Methods in Natural Language Processing}.\hskip 1em plus 0.5em minus
  0.4em\relax Austin, Texas: Association for Computational Linguistics,
  pp. 1535--1545, 2016.

\bibitem{Lample2019}
A.~Conneau and G.~Lample, ``{Cross-lingual Language Model Pretraining},''
  in \emph{Advances in Neural Information Processing Systems}, H.~Wallach et al.,
  Eds., \hskip 1em plus 0.5em minus 0.4em\relax Curran Associates, Inc., vol.~32,
  2019.

\bibitem{Gulcehre2017}
C.~Gulcehre, O.~Firat, K.~Xu, K.~Cho, and Y.~Bengio, ``{On integrating a
  language model into neural machine translation},'' \emph{Computer Speech {\&}
  Language}, vol.~45, pp. 137--148, 2017.

\bibitem{Graca2019}
M.~Gra{\c{c}}a, Y.~Kim, J.~Schamper, S.~Khadivi, and H.~Ney, ``{Generalizing
  Back-Translation in Neural Machine Translation},'' in \emph{Proceedings of
  the Fourth Conference on Machine Translation (Volume 1: Research
  Papers)}.\hskip 1em plus 0.5em minus 0.4em\relax Florence, Italy: Association
  for Computational Linguistics, pp. 45--52, 2019.

\bibitem{Poncelas2018}
A.~Poncelas, D.~Shterionov, A.~Way, G.~W. {Maillette de Buy}, and P.~Passban,
  ``{Investigating Backtranslation in Neural Machine Translation},''
  \emph{CoRR}, vol. abs/1804.0, 2018.

\bibitem{Poncelas2019}
A.~Poncelas, G.~W. {Maillette de Buy}, and A.~Way, ``{Adaptation of Machine
  Translation Models with Back-translated Data using Transductive Data
  Selection Methods},'' \emph{CoRR}, vol. abs/1906.0, 2019.

\bibitem{Poncelas2019a}
A.~Poncelas and A.~Way, ``{Selecting Artificially-Generated Sentences for
  Fine-Tuning Neural Machine Translation},'' in \emph{Proceedings of the 12th
  International Conference on Natural Language Generation}.\hskip 1em plus
  0.5em minus 0.4em\relax Association for Computational Linguistics, pp.
  219--228, 2019.

\bibitem{Adusumilli2007}
K.~K. Adusumilli, ``{Natural Languages Translation Using an Intermediate
  Language},'' \emph{IAENG International Journal of Computer Science}, vol.~33,
  no.~1, pp. 123--125, 2007.

\bibitem{Lioutas2020}
V.~Lioutas and Y.~Guo, ``{Time-aware Large Kernel Convolutions},'' in
  \emph{Proceedings of the 37th International Conference on Machine Learning},
  H.~I. Daum{\'{e}} and A.~Singh, Eds.\hskip 1em plus 0.5em minus 0.4em\relax
  PMLR, pp. 6172--6183, 2020.

\bibitem{Sennrich2016a}
R.~Sennrich, B.~Haddow, and A.~Birch, ``{Improving Neural Machine Translation
  Models with Monolingual Data},'' in \emph{Proceedings of the 54th Annual
  Meeting of the Association for Computational Linguistics}.\hskip 1em plus
  0.5em minus 0.4em\relax Berlin, Germany: Association for Computational
  Linguistics, pp. 86--96, 2016.

\bibitem{Burlot2018}
F.~Burlot and F.~Yvon, ``{Using Monolingual Data in Neural Machine Translation:
  a Systematic Study},'' in \emph{Proceedings of the Third Conference on
  Machine Translation: Research Papers}.\hskip 1em plus 0.5em minus 0.4em\relax
  Brussels, Belgium: Association for Computational Linguistics, pp.
  144--155, 2018.

\bibitem{Cotterell2018}
R.~Cotterell and J.~Kreutzer, ``{Explaining and Generalizing Back-Translation
  through Wake-Sleep},'' \emph{arXiv:1806.04402v1 [cs.CL]}, 2018.

\bibitem{Dabre2019}
R.~Dabre, K.~Chen, B.~Marie, R.~Wang, A.~Fujita, M.~Utiyama, and E.~Sumita,
  ``{NICT's Supervised Neural Machine Translation Systems for the WMT19 News
  Translation Task},'' in \emph{Proceedings of the Fourth Conference on Machine
  Translation (WMT), vol. 2: Shared Task Papers (Day 1)}, Florence, Italy, 
  pp. 168--174, 2019.

\bibitem{Kocmi2019}
T.~Kocmi and O.~Bojar, ``{CUNI Submission for Low-Resource Languages in WMT
  News 2019},'' in \emph{Proceedings of the Fourth Conference on Machine
  Translation (WMT), vol. 2: Shared Task Papers (Day 1)}, Florence, Italy,
  pp. 234--240, 2019.

\bibitem{Luo2020}
G.-x. Luo, Y.-t. Yang, R.~Dong, Y.-h. Chen, and W.-b. Zhang, ``{A Joint
  Back-Translation and Transfer Learning Method for Low-Resource Neural Machine
  Translation},'' \emph{Mathematical Problems in Engineering}, vol. 2020, pp.
  1--11, 2020.

\bibitem{Abdulmumin2021}
I.~Abdulmumin, B.~S. Galadanci, and A.~Isa, ``{Enhanced Back-Translation for
  Low Resource Neural Machine Translation Using Self-training},'' in
  \emph{Information and Communication Technology and Applications}, ccis
  1350~ed., S.~Misra and B.~Muhammad-Bello, Eds.\hskip 1em plus 0.5em minus
  0.4em\relax Minna, Nigeria: Springer Nature Switzerland AG, ch.~28, pp.
  355--371, 2021.

\bibitem{Niu2018}
X.~Niu, M.~Denkowski, and M.~Carpuat, ``{Bi-Directional Neural Machine
  Translation with Synthetic Parallel Data},'' in \emph{Proceedings of the 2nd
  Workshop on Neural Machine Translation and Generation}.\hskip 1em plus 0.5em
  minus 0.4em\relax Melbourne, Australia: Association for Computational
  Linguistics, pp. 84--91, 2018.

\bibitem{Imamura2018}
K.~Imamura, A.~Fujita, and E.~Sumita, ``{Enhancement of Encoder and Attention
  Using Target Monolingual Corpora in Neural Machine Translation},'' in
  \emph{Proceedings of the 2nd Workshop on Neural Machine Translation and
  Generation}.\hskip 1em plus 0.5em minus 0.4em\relax Melbourne, Australia:
  Association for Computational Linguistics, pp. 55--63, 2018.

\bibitem{Sennrich2016}
R.~Sennrich, B.~Haddow, and A.~Birch, ``{Edinburgh Neural Machine Translation
  Systems for WMT 16},'' in \emph{Proceedings of the First Conference on
  Machine Translation: Volume 2, Shared Task Papers}.\hskip 1em plus 0.5em
  minus 0.4em\relax Berlin, Germany: Association for Computational Linguistics,
  pp. 371--376, 2016.

\bibitem{Caswell2019}
I.~Caswell, C.~Chelba, and D.~Grangier, ``{Tagged Back-Translation},'' in
  \emph{Proceedings of the Fourth Conference on Machine Translation (Volume 1:
  Research Papers)}.\hskip 1em plus 0.5em minus 0.4em\relax Florence, Italy:
  Association for Computational Linguistics, pp. 53--63, 2019.

\bibitem{Yang2019}
Z.~Yang, W.~Chen, F.~Wang, and B.~Xu, ``{Effectively training neural machine
  translation models with monolingual data},'' \emph{Neurocomputing}, vol. 333,
  pp. 240--247, 2019.

\bibitem{Abdulmumin2019a}
I.~Abdulmumin, B.~S. Galadanci, and A.~Garba, ``{Tag-less Back-Translation},''
  \emph{arXiv:1912.10514 [cs.CL]}, 2019.

\bibitem{Fadaee2018}
M.~Fadaee and C.~Monz, ``{Back-Translation Sampling by Targeting Difficult
  Words in Neural Machine Translation},'' in \emph{Proceedings of the 2018
  Conference on Empirical Methods in Natural Language Processing}.\hskip 1em
  plus 0.5em minus 0.4em\relax Brussels, Belgium: Association for Computational
  Linguistics, pp. 436--446, 2018.

\bibitem{Kocmi2018}
T.~Kocmi and O.~Bojar, ``{Trivial Transfer Learning for Low-Resource Neural
  Machine Translation},'' in \emph{Proceedings of the Third Conference on
  Machine Translation (WMT)}, vol.~1, Brussels, Belgium, pp. 244--252, 2018.

\bibitem{Chen2017}
Z.~Chen, Y.~Tan, C.~Zhang, Q.~Xiang, L.~Zhang, M.~Li, and M.~Wang, ``{Improving
  Machine Translation Quality Estimation with Neural Network Features},'' in
  \emph{Proceedings of the Conference on Machine Translation (WMT), vol. 2:
  Shared Task Papers}.\hskip 1em plus 0.5em minus 0.4em\relax Copenhagen,
  Denmark: Association for Computational Linguistics, pp. 551--555, 2017.

\bibitem{Papineni2002}
K.~Papineni, S.~Roukos, T.~Ward, and W.-J. Zhu, ``{Bleu: a Method for Automatic
  Evaluation of Machine Translation},'' in \emph{Linguistics, Proceedings of
  the 40th Annual Meeting of the Association for Computational}.\hskip 1em plus
  0.5em minus 0.4em\relax Philadelphia, Pennsylvania, USA: Association for
  Computational Linguistics, pp. 311--318, 2002.

\bibitem{Lavie2009}
A.~Lavie and M.~J. Denkowski, ``The meteor metric for automatic evaluation of
  machine translation,'' \emph{Machine Translation}, vol.~23, no. 2–3, p.
  105–115, 2009.

\bibitem{Qiu2020}
D.~Qiu, H.~Jiang, and S.~Chen, ``{The Integrated Evaluation Method of Machine
  Translation Quality Based on Z-numbers},'' \emph{IAENG International Journal
  of Computer Science}, vol.~47, no.~1, pp. 61--67, 2020.

\bibitem{Martins2017}
A.~F. Martins, M.~Junczys-Dowmunt, F.~N. Kepler, R.~Astudillo, C.~Hokamp, and
  R.~Grundkiewicz, ``{Pushing the Limits of Translation Quality Estimation},''
  in \emph{Transactions of the Association for Computational Linguistics},
  vol.~5, pp. 205--218, 2017.

\bibitem{Specia2013}
L.~Specia, K.~Shah, J.~G.~C. de~Souza, and T.~Cohn, ``{QuEst - A translation
  quality estimation framework},'' in \emph{Proceedings of the 51st ACL: System
  Demonstrations}.\hskip 1em plus 0.5em minus 0.4em\relax Association for
  Computational Linguistics, pp. 79--84, 2013.

\bibitem{Ueffing2003}
N.~Ueffing, K.~Macherey, and H.~Ney, ``Confidence measures for statistical
  machine translation,'' in \emph{In Proc. MT Summit IX}.\hskip 1em plus 0.5em
  minus 0.4em\relax Springer-Verlag, pp. 394--401, 2003.

\bibitem{Blatz2004}
J.~Blatz, E.~Fitzgerald, G.~Foster, S.~Gandrabur, C.~Goutte, A.~Kulesza,
  A.~Sanchis, and N.~Ueffing, ``{Confidence Estimation for Machine
  Translation},'' in \emph{{\{}COLING{\}} 2004: Proceedings of the 20th
  International Conference on Computational Linguistics}.\hskip 1em plus 0.5em
  minus 0.4em\relax Geneva, Switzerland: COLING, pp. 315--321, 2004.

\bibitem{Kim2017}
H.~Kim, J.-H. Lee, and S.-H. Na, ``{Predictor-Estimator using Multilevel Task
  Learning with Stack Propagation for Neural Quality Estimation},'' in
  \emph{Proceedings of the Conference on Machine Translation (WMT), vol. 2:
  Shared Task Papers}.\hskip 1em plus 0.5em minus 0.4em\relax Copenhagen:
  Association for Computational Linguistics, pp. 562--568, 2017.

\bibitem{Kepler2019}
F.~Kepler, J.~Tr{\'{e}}nous, M.~Treviso, M.~Vera, and A.~F. Martins,
  ``{OpenKiwi: An open source framework for quality estimation},'' in
  \emph{Proceedings of the 57th Annual Meeting of the Association for
  Computational Linguistics: System Demonstrations}.\hskip 1em plus 0.5em minus
  0.4em\relax Florence, Italy: Association for Computational Linguistics,
  pp. 117--122, 2019.

\bibitem{Wang2018}
J.~Wang, K.~Fan, B.~Li, F.~Zhou, B.~Chen, Y.~Shi, and L.~Si, ``{Alibaba
  Submission for WMT18 Quality Estimation Task},'' in \emph{Proceedings of the
  Third Conference on Machine Translation (WMT), vol. 2: Shared Task
  Papers}.\hskip 1em plus 0.5em minus 0.4em\relax Brussels, Belgium:
  Association for Computational Linguistics, pp. 822--828, 2018.

\bibitem{Luong2015}
M.~T. Luong, H.~Pham, and C.~D. Manning, ``{Effective Approaches to
  Attention-based Neural Machine Translation},'' in \emph{Proceedings of the
  2015 Conference on Empirical Methods in Natural Language Processing}.\hskip
  1em plus 0.5em minus 0.4em\relax Lisbon, Portugal: Association for
  Computational Linguistics, pp. 1412--1421, 2015.

\bibitem{Sutskever2014}
I.~Sutskever, O.~Vinyals, and Q.~V. Le, ``{Sequence to Sequence Learning with
  Neural Networks},'' in \emph{Advances in Neural Information Processing Systems},
  H.~Wallach et al., Eds., \hskip 1em plus 0.5em minus 0.4em\relax Curran Associates,
  Inc., vol.~32, 2014.

\bibitem{Hochreiter1997}
S.~Hochreiter and J.~Schmidhuber, ``{Long Short-Term Memory},'' \emph{Neural
  Computation}, vol.~9, no.~8, pp. 1735--1780, 1997.

\bibitem{Dehghani2019}
M.~Dehghani, S.~Gouws, O.~Vinyals, J.~Uszkoreit, and {\L}.~Kaiser, ``{Universal
  Transformers},'' \emph{7th International Conference on Learning Representations,
  {\{}ICLR{\}}}, New Orleans, LA, USA, 2019.

\bibitem{Wu2019}
F.~Wu, A.~Fan, A.~Baevski, Y.~N. Dauphin, and M.~Auli, ``{Pay Less Attention
  with Lightweight and Dynamic Convolutions},'' in \emph{7th International
  Conference on Learning Representations, {\{}ICLR{\}}}, New Orleans, LA, USA, 2019.

\bibitem{Cho2014}
K.~Cho, B.~van Merrienboer, {\c{C}}.~G{\"{u}}l{\c{c}}ehre, F.~Bougares,
  H.~Schwenk, and Y.~Bengio, ``Learning Phrase Representations using {RNN} Encoder
  {--}Decoder for Statistical Machine Translation'' in \emph{Proceedings of the 2014
  Conference on Empirical Methods in Natural Language Processing}.\hskip 1em plus
  0.5em minus 0.4em\relax Doha, Qatar: Association for Computational Linguistics,
  pp. 1724--1734, 2014.

\bibitem{Abadi2015}
M.~Abadi, A.~Agarwal, P.~Barham, E.~Brevdo, Z.~Chen, C.~Citro, G.~S. Corrado,
  A.~Davis, J.~Dean, M.~Devin, S.~Ghemawat, I.~Goodfellow, A.~Harp, G.~Irving,
  M.~Isard, Y.~Jia, R.~Jozefowicz, L.~Kaiser, M.~Kudlur, J.~Levenberg,
  D.~Man\'{e}, R.~Monga, S.~Moore, D.~Murray, C.~Olah, M.~Schuster, J.~Shlens,
  B.~Steiner, I.~Sutskever, K.~Talwar, P.~Tucker, V.~Vanhoucke, V.~Vasudevan,
  F.~Vi\'{e}gas, O.~Vinyals, P.~Warden, M.~Wattenberg, M.~Wicke, Y.~Yu, and
  X.~Zheng, ``{TensorFlow}: Large-scale machine learning on heterogeneous
  systems,'' 2015.

\bibitem{Klein2017}
G.~Klein, Y.~Kim, Y.~Deng, J.~Senellart, and A.~M. Rush, ``{OpenNMT:
  Open-Source Toolkit for Neural Machine Translation},'' \emph{Proceedings of
  {\{}ACL{\}} 2017, System Demonstrations}, pp. 67--72, 2017.

\bibitem{Kingma2015}
D.~P. Kingma and J.~Ba, ``{Adam: A Method for Stochastic Optimization},'' in
  \emph{3rd International Conference on Learning Representations, {\{}ICLR{\}}
  2015, San Diego, CA, USA, May 7-9, 2015, Conference Track Proceedings},
  Y.~Bengio and Y.~LeCun, Eds., 2015.

\bibitem{Zhang2016a}
Y.~Zhang and D.~Weiss, ``{Stackpropagation: Improved representation learning
  for syntax},'' in \emph{Proceedings of the 54th Annual Meeting of the
  Association for Computational Linguistics (Volume 1: Long Papers)}.\hskip 1em
  plus 0.5em minus 0.4em\relax Berlin, Germany: Association for Computational
  Linguistics, pp. 1557--1566, 2016.

\bibitem{Cettolo2014}
M.~Cettolo, J.~Niehues, S.~St{\"{u}}ker, L.~Bentivogli, and M.~Federico,
  ``{Report on the 11th IWSLT Evaluation Campaign, IWSLT 2014},'' in
  \emph{Proceedings of the 11th Workshop on Spoken Language Translation}, Lake
  Tahoe, CA, USA, pp. 2--16, 2014.

\bibitem{Ranzato2016}
M.~Ranzato, S.~Chopra, M.~Auli, and W.~Zaremba, ``{Sequence Level Training with
  Recurrent Neural Networks},'' in \emph{International Conference on Learning
  Representations}, 2016.

\bibitem{Bojar2017}
O.~Bojar, R.~Chatterjee, C.~Federmann, Y.~Graham, B.~Haddow, S.~Huang, M.~Huck,
  P.~Koehn, Q.~Liu, V.~Logacheva, C.~Monz, M.~Negri, M.~Post, R.~Rubino,
  L.~Specia, and M.~Turchi, ``{Findings of the 2017 Conference on Machine
  Translation (WMT17)},'' in \emph{Proceedings of the Second Conference on
  Machine Translation, Volume 2: Shared Task Papers}.\hskip 1em plus 0.5em
  minus 0.4em\relax Copenhagen, Denmark: Association for Computational
  Linguistics, pp. 169--214, 2017.

\bibitem{Sennrich2016b}
R.~Sennrich, B.~Haddow, and A.~Birch, ``{Neural Machine Translation of Rare
  Words with Subword Units},'' in \emph{Proceedings of the 54th Annual Meeting
  of the Association for Computational Linguistics (Volume 1: Long
  Papers)}.\hskip 1em plus 0.5em minus 0.4em\relax Stroudsburg, PA, USA:
  Association for Computational Linguistics, pp. 1715--1725, 2016.

\bibitem{Koehn2004}
P.~Koehn, ``{Statistical Significance Tests for Machine Translation
  Evaluation},'' in \emph{Proceedings of the 2004 Conference on Empirical
  Methods in Natural Language Processing}.\hskip 1em plus 0.5em minus
  0.4em\relax Barcelona, Spain: Association for Computational Linguistics,
  pp. 388--395, 2004.

\bibitem{Babych2014}
B.~Babych, ``{Automated MT evaluation metrics and their limitations},''
  \emph{Tradum{\`{a}}tica: tecnologies de la traducci{\'{o}}}, no.~12, p. 464,
  2014.

\bibitem{Chatzikoumi2020}
E.~Chatzikoumi, ``{How to evaluate machine translation: A review of automated
  and human metrics},'' \emph{Natural Language Engineering}, vol.~26, no.~2,
  pp. 137--161, 2020.

\bibitem{Wiseman2016}
S.~Wiseman and A.~M. Rush, ``Sequence-to-Sequence Learning as Beam-Search
  Optimization,'' in \emph{Proceedings of the 2016 Conference on Empirical
  Methods in Natural Language Processing}.\hskip 1em plus 0.5em minus
  0.4em\relax Austin, Texas: Association for Computational Linguistics,
  pp. 1296--1306, 2016.

\bibitem{Bahdanau2017}
D.~Bahdanau, P.~Brakel, K.~Xu, A.~Goyal, R.~Lowe, J.~Pineau, A.~Courville, and
  Y.~Bengio, ``{An Actor-Critic Algorithm for Sequence Prediction},'' in
  \emph{International Conference on Learning Representations, {\{}ICLR{\}}},
  pp. 1--17, 2017.

\bibitem{Huang2018}
P.-S. Huang, C.~Wang, S.~Huang, D.~Zhou, and L.~Deng, ``{Towards Neural
  Phrase-based Machine Translation},'' \emph{International Conference on
  Learning Representations, {\{}ICLR{\}}}, 2018.

\bibitem{Shen2020}
D.~Shen, M.~Zheng, Y.~Shen, Y.~Qu, and W.~Chen, ``{A Simple but Tough-to-Beat
  Data Augmentation Approach for Natural Language Understanding and
  Generation},'' \emph{arXiv e-prints}, 2020.

\bibitem{koehn-2004-statistical}
P.~Koehn, ``Statistical significance tests for machine translation
  evaluation,'' in \emph{Proceedings of the 2004 Conference on Empirical
  Methods in Natural Language Processing}.\hskip 1em plus 0.5em minus
  0.4em\relax Barcelona, Spain: Association for Computational Linguistics,
  pp. 388--395, 2004.

\bibitem{NEURIPS2019_9015}
A.~Paszke, S.~Gross, F.~Massa, A.~Lerer, J.~Bradbury, G.~Chanan, T.~Killeen,
  Z.~Lin, N.~Gimelshein, L.~Antiga, A.~Desmaison, A.~Kopf, E.~Yang, Z.~DeVito,
  M.~Raison, A.~Tejani, S.~Chilamkurthy, B.~Steiner, L.~Fang, J.~Bai, and
  S.~Chintala, ``Pytorch: An imperative style, high-performance deep learning
  library,'' in \emph{Advances in Neural Information Processing Systems 32},
  H.~Wallach, H.~Larochelle, A.~Beygelzimer, F.~d\textquotesingle
  Alch\'{e}-Buc, E.~Fox, and R.~Garnett, Eds.\hskip 1em plus 0.5em minus
  0.4em\relax Curran Associates, Inc., pp. 8024--8035, 2019.

\end{thebibliography}

\end{document}